\documentclass{article}

\usepackage[final]{neurips_2021_ml4ad}

\usepackage[utf8]{inputenc} 
\usepackage[T1]{fontenc}    
\usepackage{hyperref}       
\usepackage{url}            
\usepackage{booktabs}       
\usepackage{amsfonts}       
\usepackage{nicefrac}       
\usepackage{microtype}      
\usepackage{xcolor}         

\usepackage{natbib}
\usepackage{tikz}
\usetikzlibrary{positioning}
\usetikzlibrary{calc}
\usepackage{subcaption}
\usepackage{amsmath}
\usepackage{algorithm}
\usepackage{algpseudocode}
\usepackage{siunitx}
\usepackage{pgfplots}
\usepackage{wrapfig}
\usepackage{amsfonts}
\usepackage{placeins}
\usepackage{makecell}

\setcitestyle{square}

\title{UMBRELLA: Uncertainty-Aware Model-Based Offline Reinforcement Learning Leveraging Planning}

\author{%
  Christopher Diehl, Timo Sievernich, Martin Krüger, Frank Hoffmann, Torsten Bertram\thanks{The authors are with the Institute of Control Theory and Systems Engineering, TU Dortmund University, D-44227, Germany.} \\
  TU Dortmund University\\
  \texttt{forename.surename@tu-dortmund.de} \\
}

\begin{document}

\maketitle

\begin{abstract}
  Offline reinforcement learning (RL) provides a framework for learning decision-making from offline data and therefore constitutes a promising approach for real-world applications as automated driving.  Self-driving vehicles (SDV) learn a policy, which potentially even outperforms the behavior in the sub-optimal data set. Especially in safety-critical applications as automated driving, explainability and transferability are key to success. This motivates the use of model-based offline RL approaches, which leverage planning. However, current state-of-the-art methods often neglect the influence of aleatoric uncertainty arising from the stochastic behavior of multi-agent systems. This work proposes a novel approach for Uncertainty-aware Model-Based Offline REinforcement Learning Leveraging plAnning (UMBRELLA), which solves the prediction, planning, and control problem of the SDV jointly in an interpretable learning-based fashion. A trained action-conditioned stochastic dynamics model captures distinctively different future evolutions of the traffic scene. 
  The analysis provides empirical evidence for the effectiveness of our approach in challenging automated driving simulations and based on a real-world public dataset. 
\end{abstract}

\section{Introduction}
Safe and interpretable motion prediction, planning, and control are crucial components for automated driving. Here, current decision-making systems are challenged by modeling multi-agent interactions. Often engineers manually design driving policies for specific scenarios, which does not scale well for more complex task. In contrast, RL \citep{sutton2018} learns a policy with little design and engineering effort but comes with the downside of trial-and-error learning. Hence, offline RL \citep{TutorialOfflineRL2020} is a promising research direction as it takes advantage of previously collected datasets and is applicable in safety-critical systems as robotics or automated driving. Moreover, in contrast to imitation learning approaches \citep{Dagger2011,Codevilla2017,Codevilla2019}, which can only recover the (sub-optimal) behavior exhibited by the dataset, offline RL methods can leverage and improve on different types of data quality. 

Applying offline RL approaches in real-world domains faces some practical challenges. Model-free \cite{CQL2020,Yin2021} and model-based policy learning methods \citep{MOReL, MOPO} have low interpretability, which is important for designing safety layers \citep{Pek2020} in automated driving. Moreover, they lack control flexibility. For example, adding new constraints often requires an expensive retraining of these model.

In contrast, the model-based offline planning (MBOP) \citep{MBOP} framework addresses the above challenges in a comprehensive manner. The algorithm utilizes different offline learned models to plan an optimal action using model-predictive control (MPC). It allows for a simple extension of the reward function and the incorporation of state constraints. Moreover, planning with the learned dynamics model enhances interpretability, which is crucial for the testing and  deployment of SDV's. While showing promising results on single-agent control tasks, MBOP only uses a simple deterministic dynamics model. However, this neglects the stochasticity of the underlying process, also known as \textit{aleatoric} uncertainty \citep{UncertaintyAleatoricEpistemic} arising from uncertainty about the behavior of other traffic agents in automated driving. In addition, MBOP operates in a fully observable setting. However, this assumption does not hold in the real-world driving setting.  For example, the intent of human drivers is not directly observable and can only be estimated indirectly from observations. 

This work tackles the previously described problems and contributes in the following way:
First, it proposes UMBRELLA, a model-based offline planning approach, which learns from offline data and plans considering both \textit{epistemic} and \textit{aleatoric} uncertainty while operating in the partially observable setting. The proposed method solves the prediction, planning, and control problem of the SDV using interpretable representations. Second, we introduce an ablation of UMBRELLA, which optimizes for the worst-case model.
Third, we demonstrate that our approach consistently outperforms behavior cloning (BC) and the state-of-the-art model-based
offline planning algorithm MBOP in urban and highway automated driving scenarios with dense traffic. 

\section{Related Work}
The proposed approach is situated within the broader literature on model-based offline RL and interaction-aware motion prediction and planning. This section provides a brief description of the underlying concepts and how they relate to the work presented here.

\textbf{Model-based Offline Reinforcement Learning.}
As in the offline RL setting, the agents do not interact with the environment during learning, one major challenge arises: the \textit{distributional shift} \cite{TutorialOfflineRL2020}. Model-based offline RL methods like MOReL and MOPO \citep{MOReL, MOPO} address this issue by incorporating an epistemic uncertainty estimate of the dynamics model into the reward function to penalize states which are not covered by the behavior distribution. \citep{yu2021combo} penalizes the value function on out-of-distribution states, making the challenging uncertainty estimation of MOReL and MOPO obsolete. MBOP uses planning based on bootstrap ensembles \citep{Bootstrap2017} of offline learned models \citep{MBOP}. MOPP improves upon MBOP by pruning trajectories to avoid potential out-of-distribution samples \citep{zhan2021modelbased}. 
\cite{Badger} introduce BADGR, a system for self-supervised robot navigation learned from off-policy data.
All these model-based approaches are evaluated in single-agent environments. Hence, they lack the modeling of aleatoric uncertainty, which is caused by uncertain human behavior in the interactive automated driving setting. \citep{MPUR} address this issue by a stochastic dynamics model represented by a conditional variational autoencoder (CVAE) \citep{Kingma2014}. However, their approach relies on policy learning, which in contrast to model-based offline planning comes with a reduced interpretability and control flexibility. 

\textbf{Interaction-aware Motion Prediction and Planning.}
Traditional automated driving pipelines follow the concept of planning a safe motion based on the prediction of all other agents, which neglects the interaction between both planning and prediction. Game-theoretic approaches \citep{ALGames,LucidGames} model the multi-agent dynamics as games but come with the challenge of computing (Nash) equilibria, and their computation times scale poorly with the number of agents. In contrast, learning-based approaches as \citep{liu2021deep,CFO2021} have the potential to generalize to a higher number of scenarios and learn from high-dimensional sensor data. However, in these works, epistemic uncertainty is not explicitly considered, which could lead to failure in out-of-distribution states.

\begin{figure}[!t]
	\begin{minipage}[b]{.45\textwidth}
		\centering
		\begin{tikzpicture}
			\node [minimum width=1cm, minimum height=1cm, fill=white, draw=black, rectangle, rounded corners=4pt, text width=2cm, align=center] (Planning) {Planning};
			
			\node [right=of Planning, minimum width=1cm, minimum height=1cm, fill=white, draw=black, rectangle, rounded corners=4pt, text width=2cm,align=center] (PolVal) {BC Policy and Value Function};
			
			\node [minimum width=1cm, minimum height=1cm, fill=white, draw=black, rectangle, rounded corners=4pt, text width=2cm, align=center] (Model) at ([yshift=2cm] $(Planning)!0.5!(PolVal)$) {Dynamics Model};
			
			\node [align=center] (act) at ([yshift=-2cm] $(Planning)!0.5!(PolVal)$) {Act};
			
			\node [align=center] (Data) at ([yshift=-2.75 cm] $(Planning)!0.5!(PolVal)$) {Data};
			
			
			\draw[ultra thick,->] (PolVal.120) --  ++(0,0.5) -| (Planning.60);
			\draw[dashed,->] (Planning.300) --  ++(0,-0.5) -| (PolVal.240);
			\draw[ultra thick,->] (Planning.240) |- (act.180);
			\draw[dashed,->] (PolVal.300) |- (act);
			\draw[ultra thick,->] (Data.0) -- ++(2.8,0) |-(PolVal.0);
			\draw[ultra thick,->] (Data.0) -- ++(2.8,0) |-(Model.0);
			\draw[ultra thick,->] (Model.180) -| (Planning.120);
			
		\end{tikzpicture}
		(a)
	\end{minipage}
	\label{UMBRELLA_Loop}
	\begin{minipage}[b]{0.52\textwidth}
		\centering
	\begin{tikzpicture}
		\node[anchor=south west,inner sep=0] (image) at (0,0) {\includegraphics[width=\textwidth]{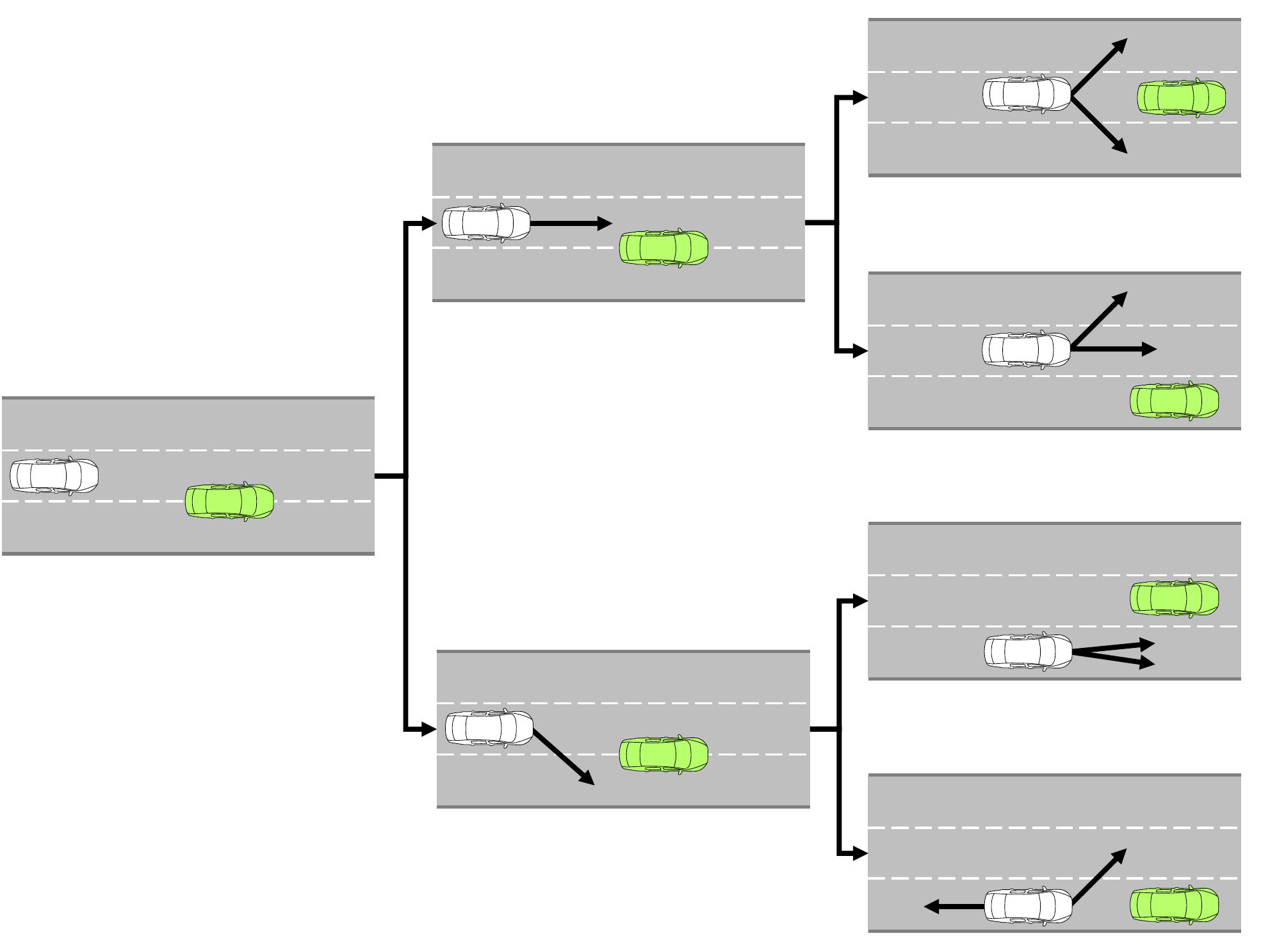}};

		\node[anchor=west](a111) at (1.8,4.25) {\scriptsize $\textbf{a}_{1}^{1}$};
		\node[anchor=west](a121) at (1.8,1.25) {\scriptsize $\textbf{a}_{1}^{2}$};
		
		\node[anchor=west](a112) at (3,4.4) {\scriptsize $\textbf{a}_{1}^{1}$};
		\node[anchor=west](a122) at (3.1,1.4) {\scriptsize $\textbf{a}_{1}^{2}$};

		\node[anchor=west](z11) at (4.3,4.9) {\scriptsize $\textbf{z}_{1}^{1}$};
		\node[anchor=west](z12) at (4.3,3.5) {\scriptsize $\textbf{z}_{1}^{2}$};
		
		\node[anchor=west](z11) at (4.3,2) {\scriptsize $\textbf{z}_{1}^{1}$};
		\node[anchor=west](z12) at (4.3,0.5) {\scriptsize $\textbf{z}_{1}^{2}$};

		\node[anchor=west](a121) at (5.75,5.2) {\scriptsize $\textbf{a}_{2}^{1}$};
		\node[anchor=west](a221) at (5.75,4.65) {\scriptsize $\textbf{a}_{2}^{2}$};
		
		\node[anchor=west](a122) at (5.75,3.75) {\scriptsize $\textbf{a}_{2}^{1}$};
		\node[anchor=west](a222) at (6.6,3.5) {\scriptsize $\textbf{a}_{2}^{2}$};
		
		\node[anchor=west](a123) at (6.1,2) {\scriptsize $\textbf{a}_{2}^{1}$};
		\node[anchor=west](a223) at (6.6,1.7) {\scriptsize $\textbf{a}_{2}^{2}$};
		
		\node[anchor=west](a124) at (4.9,0.3) {\scriptsize $\textbf{a}_{2}^{1}$};
		\node[anchor=west](a224) at (5.9,0.6) {\scriptsize $\textbf{a}_{2}^{2}$};

	\end{tikzpicture}
	(b)
	\end{minipage}

	\caption{(a) UMBRELLA's integration of learning and planning following the general visualization of \cite{moerland2021modelbased}: Model learning includes the dynamics, BC policy, and value function. The thick arrows indicate which parts of the planning/learning loop are used here. There is no arrow from \textit{Act} to \textit{Data} due to offline learning. b) One step of the planning procedure. For the sake of clarity, the number of samples equals $N=2$. Actions and latent variables are indexed by time step $t$ and sample $n$, resulting in $\textbf{a}_t^n$ and $\textbf{z}_t^n$, respectively.}
	\label{fig:UMBOP_planning}
\end{figure}
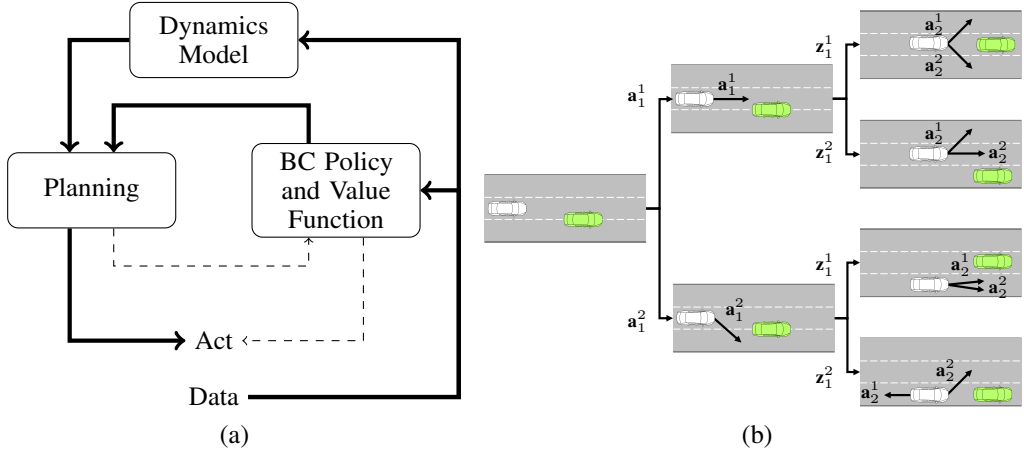

\section{The UMBRELLA Framework}
The proposed algorithm is a model-based offline RL algorithm which learns from previously recorded datasets. UMBRELLA learns a stochastic dynamics model, a BC policy, and a truncated value function as shown in Figure \ref{fig:UMBOP_planning} a). UMBRELLA is an extension of the MBOP \citep{MBOP} method and plans for different future evolutions. Each model is a bootstrap ensemble \citep{Bootstrap2017} of $K$ neural networks. The weights of the each ensemble head $f^i$ with $i \in \left[1,\dots\,K \right] $ are initialized differently but get trained on the same dataset $D$. 

\subsection{Problem Formulation}
The RL task is to control a dynamical control, described by a \textit{Markov Decision Process} (MDP). The MDP is defined by the tuple $\mathcal{M}=(\mathcal{S}, \mathcal{A}, p, r, \gamma)$, where $\mathcal{S}$ is a set of \textit{states} $\textbf{s}\in \mathcal{S}$ and $\mathcal{A}$ is a set of \textit{actions} $\textbf{a} \in \mathcal{A}$. Let $r(\textbf{s}_t, \textbf{a}_t, \textbf{s}_{t+1})$ denote the \textit{reward}, which the agent receives when it takes action $\textbf{a}_t$ in state $\textbf{s}_t$ at timestep $t$ and arrives in state $\textbf{s}_{t+1}$ with a probability described by the \textit{transition dynamics} $p(\textbf{s}_{t+1} \mid \textbf{s}_t, \textbf{a}_t)$. 
The \textit{policy} $\pi : \mathcal{S} \rightarrow \mathcal{A} $ maps from states to a probability distribution over actions. The objective is to estimate an \textit{optimal policy} function
$\pi^* = \arg \max_{a \in \mathcal{A}}\sum_{t=1}^{H}\gamma^t r(\textbf{s}_t, \textbf{a}_t, \textbf{s}_{t+1})$ that maximizes the finite-horizon cumulative reward over the horizon $H$. $\gamma = 1$ is the scalar time-wise discount factor. In the offline setting, the agent does not interact with the environment but rather learns from a static dataset $\mathcal{D}$ generated by a behavior policy $\pi_d$.  
When the sampled observations $\textbf{o}_t \in \mathcal{O}$ do not fully capture the ground truth state, the MDP becomes a partially observable MDP (POMDP). It is described by the tuple $\mathcal{M}_{PO}=(\mathcal{S}, \mathcal{A}, \mathcal{O}, p, r, \gamma)$. A common method to provide a solution in the partially observable setting is the \textit{nth-order history} \citep{sutton2018} approach. Here the state estimation is approximated by a sequence of the $n_\textrm{c}\in \mathbb{N}^+_0 $ last observations and actions.  The result is a large but finite Markov decision process (MDP) and standard RL approaches can then be applied.

\subsection{General Approach}
Forecasting other agent responses to the actions of the SDV vehicle is critical for automated driving, and faces the challenge of the inherent uncertainty of human behavior. As an example, let us consider the situation illustrated in Figure \ref{fig:UMBOP_planning} b). Here another agent (green) starts to cut in onto the lane of the SDV (white). The SDV has to estimate the probability of each future outcome (i.e. the other agent aborts or continues the maneuver) and should plan its action $\textbf{a}_t$ accordingly. UMBRELLA models the stochasticity of other agents' behavior at time $t$ with a continuous latent variable $\textbf{z}_t \in \mathcal{Z}$. As enumerating all possible actions of the SDV during planning is intractable, we sample actions based on a learned BC policy (Section \ref{BC policy}). That leads to efficient, expert-like rollouts of $N\in \mathbb{N}^+_0$ potential state-trajectories over a planning horizon $H$. After trajectory sampling, the approach applies a return-weighted trajectory optimizer. This work follows the nth-order history approach to account for states, which are not fully observable (e.g. human drivers intent) and are merely estimated from the last observations $\textbf{o}_{t-n_\textrm{c}:t}$ up to timestep $t$.  

\subsection{Stochastic Dynamics Model}
To model different futures, we learn stochastic forward dynamics models $f_{\mathrm{m},\boldsymbol{\theta}}: \mathcal{S} \times \mathcal{A} \times \mathcal{Z} \to \mathcal{S} \times \mathbb{R}$ parameterized by $\boldsymbol{\theta}$ based on the work of \citep{MPUR}.  As the evolution of the traffic scene also depends on the action $\textbf{a}_t$ of the SDV, a dynamics model is action-conditioned and answers the question: \textit{If the SDV takes this action, how will the other agents in the scene respond?} UMBRELLA's augmented models captures the interaction between prediction and planning in the context of automated driving. The model, which is a CVAE \citep{Kingma2014}, outputs the prediction for the next state $\hat{\textbf{s}}_{t+1}$ = $f_{\mathrm{m}}(\textbf{s}_t,\textbf{a}_t,\textbf{z}_t)_{\mathrm{s}}$ and the corresponding reward prediction $\hat{r}_{t}$ = $f_{\mathrm{m}}(\textbf{s}_t,\textbf{a}_t,\textbf{z}_t)_{\mathrm{r}}$. As the model outputs two predictions, we minimize a multi-task mean squared error loss during training. 

 The latent variable $\textbf{z}_t$ models different future predictions and makes sure the output is non-deterministic to the input.
 During training, the latent variable is sampled from the posterior distribution  $q_{\boldsymbol{\phi}}(\textbf{z}\mid \textbf{s}_t, \textbf{s}_{t+1})$ parameterized by $\boldsymbol{\phi}$. As we can only sample from the prior distribution during inference, the Kullback-Leibler (KL) divergence between the posterior and the prior distribution $p(\textbf{z})$ is also minimized, following the definition of the Evidence Lower BOund (ELBO) objective used when training VAEs. Using a weighting factor $\boldsymbol{\zeta}$ the per-sample loss is then given by 
 \begin{equation}
 	\begin{split}
 		\label{eq:loss_fm2}
 		\mathcal{L}(\boldsymbol{\theta}, \boldsymbol{\phi}; \textbf{s}_t, \textbf{s}_{t+1}, \textbf{a}_t, r_t) = & {|| \textbf{s}_{t+1} - f_{\mathrm{m},\boldsymbol{\theta}}(\textbf{s}_t,\textbf{a}_t,\textbf{z}_t)_{\mathrm{s}}||}_2^2  +{|| r_{t} - f_{\mathrm{m},\boldsymbol{\theta}}(\textbf{s}_t,\textbf{a}_t,\textbf{z}_t)_{\mathrm{r}}||}_2^2 + \\& + \zeta D_{\mathrm{KL}}(q_{\boldsymbol{\phi}}(\textbf{z}_t \mid \textbf{s}_t, \textbf{s}_{t+1})||p(\textbf{z}_t))
 	\end{split}
 \end{equation}

\subsection{Behavior Cloned Policy and Truncated Value Function}
\label{BC policy}
During inference, the algorithm rolls out potential state trajectories according to the stochastic forward dynamics model. For this, an action-sampling procedure guided by a BC policy is applied. UMBRELLA learns a bootstrap ensemble of BC policies $f_{\mathrm{b},\boldsymbol{\psi}}: \mathcal{S} \times \mathcal{A}^{n_c} \to \mathcal{A}$ parameterized by $\boldsymbol{\psi}\in \mathbb{R}$. The model $f_\mathrm{b}(\textbf{s}_t, \textbf{a}_{(t-n_c):(t-1)})$ takes the current state and the $n_c$ previous actions as input and outputs the action $\textbf{a}_t$.  By concatenating the previous actions the learned action $\textbf{a}_t$ is supposed to be more smoothly. 

Computational resources limit the planning horizon length. Therefore prior works \citep{POLO, MBOP, zhan2021modelbased}  successfully propose a value function to extend the planning horizon. UMBRELLA also learn a truncated value function $f_{\mathrm{R},\boldsymbol{\xi}}: \mathcal{S} \times \mathcal{A}^{n_c} \to \mathbb{R}$ parameterized by $\boldsymbol{\xi}$
to estimate the expected return of the next $H$ episodes $\hat{R}_H$ when being in state $\textbf{s}_t$ and executed $\textbf{a}_{(t-n_c):(t-1)}$ as $n_c$ previously taken actions. That effectively extends the planning horizon without increasing the number of rollouts of the dynamics model.
Appendix \ref{Model Architecture} and \ref{Training Details} report additional details of the models' architecture and training procedure.

\subsection{UMBRELLA-Planning}
\textbf{Model Predictive Control Formulation.} UMBRELLA uses model-predictive control (MPC) \citep{MPC}, which has a long history in control and automation, to plan it's actions. Hence in every planning step, the algorithm solves a finite-horizon optimal control problem resulting in an optimal trajectory $T$ of length $H$. Then the first action of the optimal control sequence is executed. The repetitive solution of an optimal control problem reduces the influence of modeling errors. 

\textbf{UMBRELLA Trajectory Optimizer.} Next, we will describe the UMBRELLA planning algorithm \ref{alg:UMBOP_planning}. It is used in every MPC planning cycle to receive an optimal action trajectory. Note that for the sake of clarity, we omit indexing of the models by the parameters $\boldsymbol{\psi},\boldsymbol{\phi},\boldsymbol{\xi}$. The algorithm generates $N$ trajectories, whereas $M\in \mathbb{N}^+_0$ trajectories are planned per ensemble head using the sampled latent variables $\textbf{z}_m$ from the prior distribution with $m \in \left[1,\dots\,M \right]$. As human drivers are assumed to not switch their driving style erratically, the latent variable is fixed across the entire trajectory indexed by $n \in \left[1,\dots,N\right]$ (Line 6), which results in a consistent prediction. Moreover, as \citep{nagabandi2020} and \citep{MBOP}, the same  $l^\textrm{th}$ ensemble head of the BC policy $f^l_\textrm{b}$ and the dynamics model $f^l_\textrm{m}$ are employed consistently across the entire trajectory. 

The BC policy guides the expansion of the trajectory by sampling an action $\textbf{a}_t$ from the BC policy with added Gaussian noise (Line 14). Afterward, the action is averaged together with the trajectory of the previous time-step (Line 15) using the mixture coefficient $\beta$ \citep{MBOP}. UMBRELLA then rolls out state-trajectories using the dynamics model $f^l_m$ (line 16) and calculates the average reward over all ensemble members (line 17). At the end of the trajectory, it calculates an average over all ensemble members of the truncated value function. Averaging over all models of the same ensemble family is inspired by prior work \citep{nagabandi2020, MBOP}. 

Last, an optimal action-trajectory is calculated using the model predictive path integral (MPPI) framework \citep{MPPI2017}. Prior works also employ MPPI in the online and offline planning setting \citep{nagabandi2020,MBOP,Badger}. Let $\textbf{A}_{N,H}$ denote set the of $N$ generated action trajectories and $\textbf{R}_N$ the corresponding rewards. Then the optimal trajectory is obtained by re-weighting each action based on the associated reward: 
\begin{equation}
\textbf{T}^*_t = \frac{\sum_{n=1}^{N}{e^{\kappa \textbf{R}_n}\textbf{A}_{n,t+1}}}{\sum_{n=1}^{N}{e^{\kappa \textbf{R}_n}}},\forall t \in [0,\dots,H-1]
\label{equ::MPPI}
\end{equation}

\begin{algorithm}[tbh]
	\caption{UMBRELLA Planning}
	\label{alg:UMBOP_planning}
	\begin{algorithmic}[1]
		\Procedure{UMBRELLA-Planning}{$\textbf{s},\textbf{T}, f_\mathrm{m}, f_\mathrm{b}, f_\mathrm{R}, H, N, \sigma^2, \beta, \kappa$} 
		\State Set $\textbf{R}_N = \vec{0}_N, \textbf{A}_{N,H} = \vec{0}_{N,H} $  \Comment{Returns and actions of $N$ sampled trajectories.}
		\State $M=N/K$ \Comment{Plan $M$ different trajectories per head.}
		\For{$m=1,\dots,M$}
		\State $\textbf{z}_m \sim p(\textbf{z})$ \Comment{Sample latent variable $\textbf{z}_m$ from prior $p(\textbf{z})$.}
		\EndFor
		\For{$n=1,\dots,N$} \Comment{Sample $N$ trajectories over horizon $H$}
		\State $l=n \mod K$ \Comment{Use $K$ different ensemble heads.}
		\State $m=n \mod M$ 
		\State $\textbf{s}_1=\textbf{s}, \textbf{a}_0=\textbf{T}_0, R=0$ 
		\For{$t=1,\dots,H$}
		\State $\boldsymbol{\epsilon} \sim \mathcal{N}{(0, \sigma ^{2})}$
		\State $\textbf{a}_t=f_{\mathrm{b}}^{l}(\textbf{s}_t,\textbf{a}_{(t-n_c):(t-1)})+\boldsymbol{\epsilon}$ \Comment{BC policy prior to guide the action sampling.}
		\State $\textbf{A}_{n,t}=(1-\beta)\textbf{a}_t+\beta \textbf{T}_{i=\min(t,H-1)}$ \Comment{Beta-mixture with prev. trajectory $\textbf{T}$.}
		\State $\textbf{s}_{t+1}=f_{\mathrm{m}}^{l}(\textbf{s}_t,\textbf{A}_{n,t},\textbf{z}_m)_{\textrm{s}}$ \Comment{Sample next state from stochastic dyn. model.}
		\State $R=R+\frac{1}{K}\sum_{i=1}^{K}{f_{\mathrm{m}}^{i}(\textbf{s}_t,\textbf{A}_{n,t},\textbf{z}_m)_{\textrm{r}}}$ \Comment{Take avg. reward over heads.}
		\EndFor
		\State $\textbf{R}_n=R+\frac{1}{K}\sum_{i=1}^{K}{f_{\mathrm{R}}^{i}(\textbf{s}_{H+1},\textbf{A}_{n,(H-n_c+1):(H)})}$ \Comment{Append pred. truncated value.}
		\EndFor
		\State Compute $\textbf{T}^*_t$ according to Eq. (\ref{equ::MPPI})
		\Comment{Compute return-weighted trajectory}
		\State \Return $\textbf{T}^*_t$
		\EndProcedure
	\end{algorithmic}
\end{algorithm}

\textbf{Pessimistic Trajectory Optimizer.} This work also proposes UMBRELLA-P, a pessimistic trajectory optimizer. UMBRELLA and MBOP \citep{MBOP} use all sampled trajectories for the computation (equation \ref{equ::MPPI}) of the weighted trajectory. UMBRELLA-P only aggregates those trajectories of the ensemble head with the lowest sum over returns. Therefore, the algorithm first calculates $\textbf{R}_{k,\textrm{sum}} \forall k \in \left[1,\dots\,K \right]$, which is the sum over all rewards belonging to the ensemble indexed by $k$. Then, it picks the index of the ensemble with the lowest sum over returns by $k^*= \arg\min_{k \in \left[1,\dots\,K \right]} \textbf{R}_{k,\textrm{sum}}$. Lastly, equation \ref{equ::MPPI} is only applied to the trajectories belonging to $k^*$.  Therefore, UMBRELLA-P optimizes in the face of epistemic uncertainty for the worst-case outcome and acts pessimistically.


\section{Experimental Evaluation}
This section evaluates the proposed method in two interactive automated driving environments. The goal of this section is to answer the following research questions:
\textit{Q1}: Does the modeling of aleatoric uncertainty enhance the performance of model-based offline planning methods in interactive automated driving scenarios?
\textit{Q2}: Does the algorithm improve upon simple behavior cloned policies?
\textit{Q3}: Does the pessimistic variant UMBRELLA-P improve the planning performance?   
\textit{Q4}: What are the limitations of model-based offline planning methods in the context of automated driving? 

\textbf{Environments.} \textit{NGSIM}: The first environment is a challenging multi-agent automated driving environment from the work of \citep{MPUR} based on the Next Generation Simulation program's Interstate 80 (NGSIM I-80) dataset \citep{NGSIM}. The goal of the SDV is to stay in the middle of a lane while avoiding collisions. \textit{CARLA}: The second environment includes a urban multi-agent scenario implemented in the CARLA simulator \citep{CARLA}. The SDV is supposed to make progress along the route and execute an unprotected left turn at a busy intersection while avoiding collisions.  Figure \ref{fig:planned_traj} (a)-(c) illustrates both experiments.  

\textbf{Baselines.} This work benchmarks against the following methods: 
(i) \textit{1-step IL}: A learned policy imitating the expert driver using the BC policy of Section \ref{BC policy} (ii) \textit{MBOP} \citep{MBOP}: A current state-of-the-art model-based offline RL method that uses the deterministic dynamics model of \cite{MPUR}. Note, that for a fair comparison all other components of MBOP are identical to the UMBRELLA approach.
In the NGSIM environment we further compare against (iii) \textit{MPUR}: \citep{MPUR} A state-of-the-art model-based policy learning method, which accounts for epistemic and aleatoric uncertainty (iv) \textit{Human}: The ground truth actions of the human.
(v) \textit{No action}: A policy that always applies actions of zeros.

\textbf{Metrics.} The comparative analysis uses the following metrics. \textit{Success rate} (SR): The rate of collision-free episodes. In the CARLA experiment, also an episode is considered as failure when the vehicle does not reach the target location within the allocated time.  \textit{Mean distance} (MD): The distance traveled longitudinal direction along the NGSIM-highway averaged over all episodes. The NGSIM analysis further evaluated the individual parts of the reward function (i.e. \textit{mean proximity reward} $\bar{r}_{\textrm{prox}}$ and \textit{mean lane reward} $\bar{r}_{\textrm{lane}}$ and the mean final reward $\bar{r}$. The CARLA experiment determines the \textit{mean successful time} (MST): The average time required for successfully navigating through the intersection averaged over all successful episodes.
Appendix \ref{Experimental Setup} provides further information about the state, action, and reward representation.

\subsection{NGSIM Experiments}
The NGSIM I-80 dataset consists of 45 minutes of real-world driving data on a highway with dense traffic recorded with cameras. The dataset has high environment (or aleatoric) uncertainty as all agents exhibit different driving maneuvers as lane changes, merges, braking, and acceleration maneuvers.  We follow the preprocessing and experimental setup of \cite{MPUR} utilizing the OpenAI Gym \citep{OpenAIGym}. In each evaluation episode, the SDV is initialized at the beginning of a random trajectory of the test set. Other agents drive along their fixed trajectories from the dataset. Fixed trajectories avoid the need of hand-designing (unrealistic) driving policies of neighboring cars. However, this log-replay comes with the downside that the other agents do not react to the SDV, which enhances the difficulty of the task as multiple agents can trap the SDV. 

\begin{figure}[!t]
	\centering
	\captionsetup[subfigure]{labelformat=empty}
	\begin{tabular}{c}
		\begin{subfigure}[t]{0.225\textwidth}
			\centering
			\begin{tikzpicture}
				
				\node[anchor=south west,inner sep=0] (image) at (0,0) {\includegraphics[width=\textwidth]{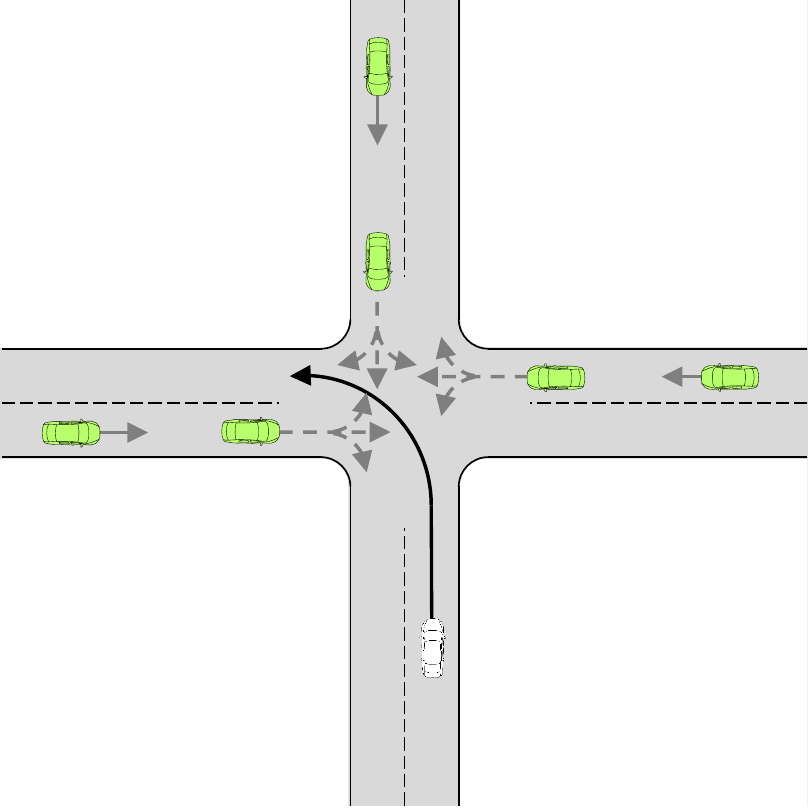}};
			\end{tikzpicture}
			\caption{(a)}
			\label{fig:CARLA_schema}
		\end{subfigure}
		\hfill
		\begin{subfigure}[t]{0.225\textwidth}
			\centering
			\begin{tikzpicture}
				
				\node[anchor=south west,inner sep=0] (image) at (0,0) {\includegraphics[width=\textwidth]{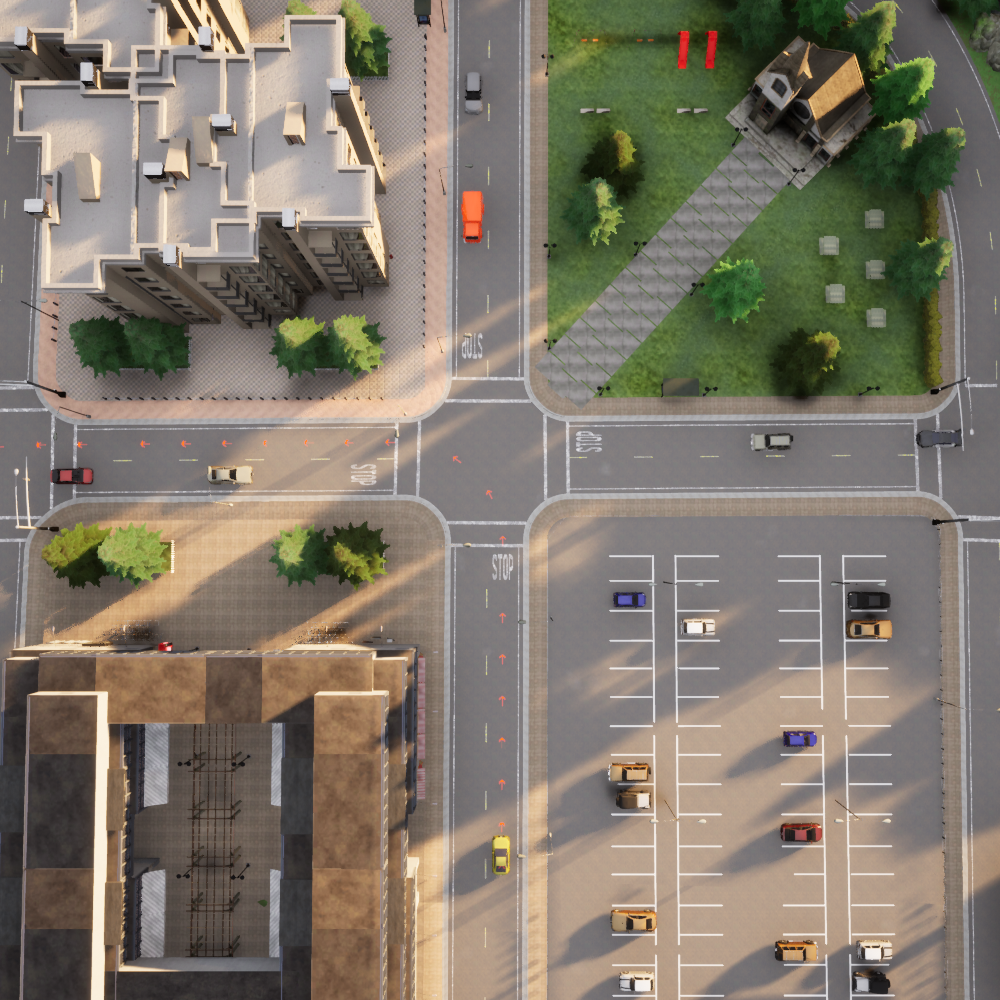}};
				\node[circle, draw=white, dashed, very thick, minimum size=0.7cm] (circle) at (3.76,1.1) {};
			\end{tikzpicture}
			
			\caption{(b)}
			\label{fig:CARLA_bev}
		\end{subfigure} 
		\\
		\begin{subfigure}{0.45\textwidth}
			\begin{tikzpicture}

				\node[anchor=south west,inner sep=0] (image) at (0,0) {\includegraphics[width=\textwidth]{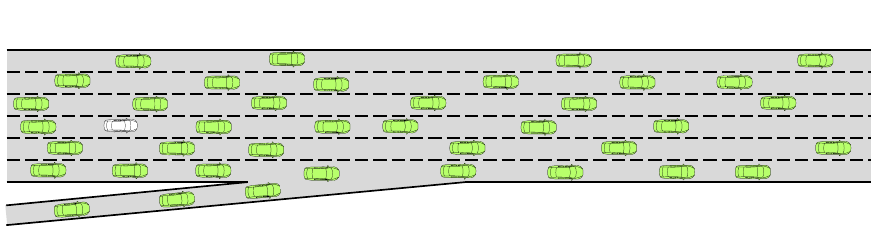}};

			\end{tikzpicture}
			\caption{(c)}
		\end{subfigure}
	\end{tabular}
	\hfill
	\resizebox{0.5\textwidth}{!}{
	\begin{subfigure}{0.1\linewidth}
		\includegraphics[width=\linewidth]{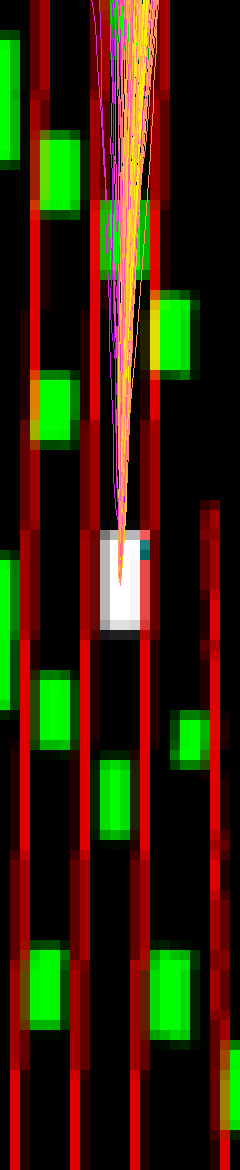}
		\caption{\scalebox{1.3}{\phantom{(a)}}}
	\end{subfigure}
	\hfill
	\begin{subfigure}{0.1\linewidth}
		\includegraphics[width=\linewidth]{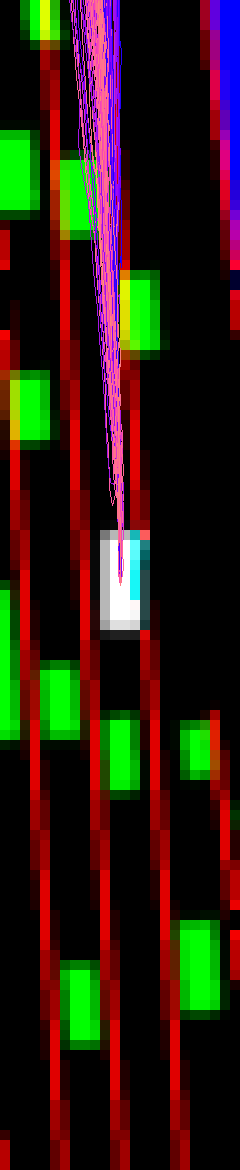}
		\caption{\scalebox{1.3}{\vphantom{(d)}}}

	\end{subfigure}
	\hfill

	\begin{subfigure}{0.1\linewidth}
		\includegraphics[width=\linewidth]{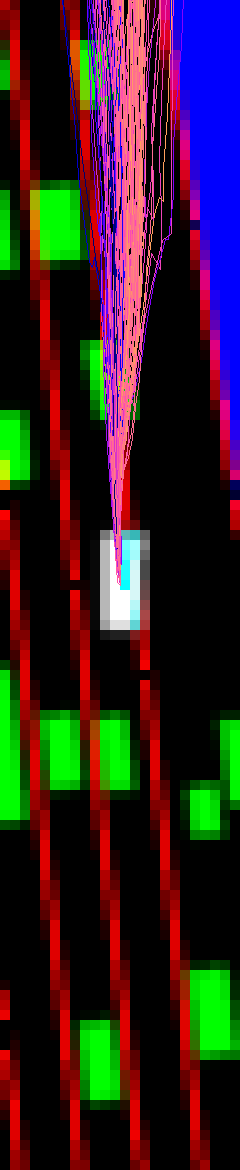}
		\caption{\scalebox{1.3}{(d)}} 
	\end{subfigure}
	\hfill

	\begin{subfigure}{0.1\linewidth}
		\includegraphics[width=\linewidth]{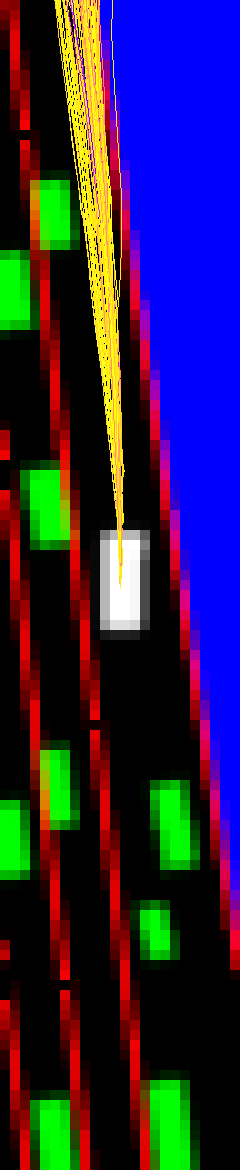}
		\caption{\scalebox{1.3}{\vphantom{(d)}}}
	\end{subfigure}
	\hfill
	\begin{subfigure}{0.1\linewidth}
		\includegraphics[width=\linewidth]{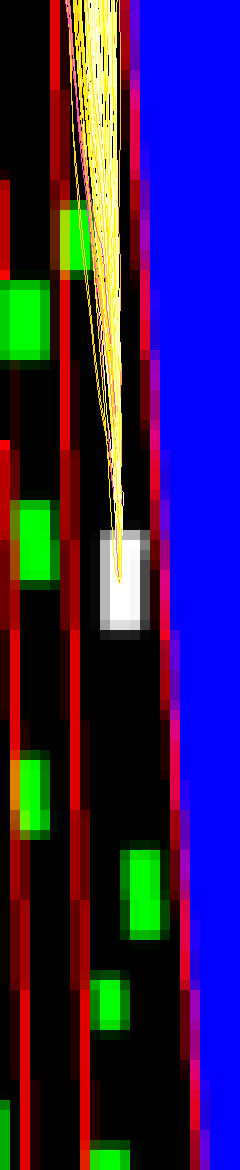}
		\caption{\scalebox{1.3}{\vphantom{(d)}}}
	\end{subfigure}
	\hfill

	\begin{subfigure}{0.0665\linewidth}
		\begin{tikzpicture}
			\node[anchor=south west,inner sep=0] (image) at (-0.5,1.2) {\includegraphics[width=\textwidth]{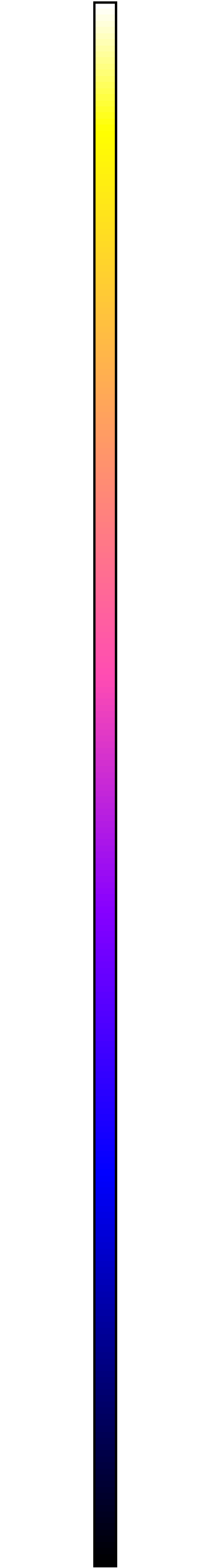}};
			
			\node[rotate=90](ret) at (0.2,4.5){\footnotesize Predicted Return};
			\draw[->] (0.2,6.8) -- (0.2,8) node[midway,below,rotate=90]{\footnotesize high};
			\draw[->] (0.2,2.6) -- (0.2,1.2) node[midway,below,rotate=90]{\footnotesize low};
			
		\end{tikzpicture}
	\caption{\scalebox{1.3}{\vphantom{(d)}}}
		\end{subfigure}
	}
\caption{(a) Schematic representation of the left turn task. (b) Intersection in CARLA. (c) NGSIM environment.  (d) Context images overlaid with the planned UMBRELLA trajectories in the time interval $t\in \left[0\si{\second},8\si{\second}\right]$ in steps of $\num{2} \si{\second}$ during a lane change maneuver. The trajectories are colored concerning their predicted return colors. The background describes the context image: The SDV is denoted in white, other agents are visualized in green. Black denotes the road and red lane markings. }
\label{fig:planned_traj}
\vspace{-0.5cm}
\end{figure}
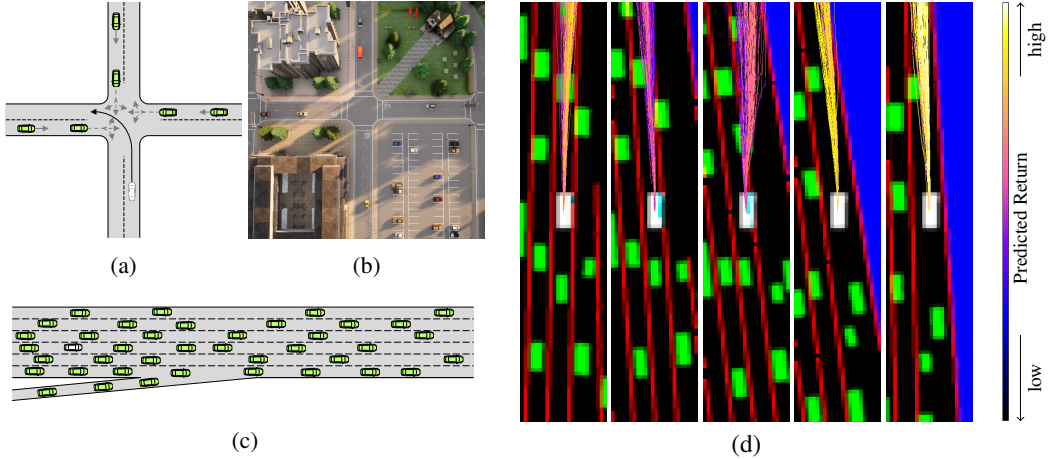

 UMBRELLA is compared against the previously described baselines. Table \ref{tab:eval} illustrates the results. In terms of success rate and mean traveled distance, MPUR performs best followed by UMBRELLA. However, MPUR does not exhibit the control flexibility as UMBRELLA and requires retraining whenever new parts are added to the reward function. In contrast, model-based offline planning methods like MBOP and UMBRELLA can easily be modified, which was shown in prior work \citep{MBOP}. Moreover, in contrast to MPUR, these methods do not require a differentiable reward function.

Regarding questions \textit{Q1} and \textit{Q2}, UMBRELLA outperforms both MBOP and the 1-step IL method. We can conclude that modeling aleatoric uncertainty is critical when applying model-based offline RL approaches in automated driving scenarios. Prediction with a deterministic dynamics model leads to an averaging over multiple futures. Hence, UMBRELLA outperforms MBOP w.r.t the considered metrics. Interestingly, within the IL approach the SDV fails on every episode. However, UMBRELLA improves drastically upon this policy. Considering question \textit{Q3}, UMBRELLA-P is inferior compared to UMBRELLA on the metrics SR and MD, while achieving slightly higher mean rewards. Our experiments revealed, that UMBRELLA-P performs fewer lane changes and it gets more frequently trapped by the non-reactive agents. We hypothesize that this occurs due to the more conservative planning. As the pessimistic variant plans for the worst-case, it stays on the same lane most of the time to avoid potential risk of lane changes. As a result, the follower vehicle causes a rear collision, as its original trajectory is ignorant of the phantom SDV.

Figure \ref{fig:planned_traj} (d) provides an exemplary visualization of the planning result in a challenging lane change scenario with dense traffic. The SDV predicts a safety-critical future evolution of the scenario (visualized with low return trajectories) as the follower vehicle starts to accelerate and potentially traps the SDV. However, the SDV reacts accordingly and successfully changes its lane in dense traffic.

\begin{table}[!t]
	\centering
	\caption{Performance of the different evaluated methods on the highway (NGSIM) and urban (CARLA) control task by diverse metrics. CARLA (ID) denotes evaluation using simulations with parameters from the same distribution as used during training. CARLA (OOD) defines experiments, where a distribution shift is simulated. The unit of the metric MD is meter. The values of the MST metric are denoted in seconds. Bold numbers mark the best result and an underlined number the second best.}
	\resizebox{\textwidth}{!}{
	\begin{tabular}{l c c c c c c c c c }
		\toprule
		& \multicolumn{5}{c}{NGSIM} & \multicolumn{2}{c}{CARLA ID} &\multicolumn{2}{c}{CARLA OOD} \\
		\cmidrule(r){2-6}
		\cmidrule(r){7-8}
		\cmidrule(r){9-10}
		Method & SR & MD & $\bar{r}_{\textrm{prox}}$ & $\bar{r}_{\textrm{lane}}$ & $\bar{r}$ & SR & MST & SR & MST  \\
		\midrule
		UMBRELLA & $\underbar{0.60}$ & $\underbar{320.06}$ &  $0.817 $ & $\underbar{0.902}$ & $1.718$ & $\textbf{0.63}$ & $\textbf{11.62}$ & $\underbar{0.43}$ & $\textbf{13.10}$ \\
		UMBRELLA-P & $0.53$ & $300.43$ & $\underbar{0.820}$ & $\textbf{0.904}$ & $\underbar{1.724}$ & $\underbar{0.50}$ &$11.97$ & $\textbf{0.53}$ & $13.55$\\
		\midrule
		MBOP & $0.50$ & $290.62$ & $0.775$ & $0.893$ & $1.669$ & $0.27$ & $\underbar{11.78}$ & $0.20$ & $\underbar{13.37}$   \\
		1-step IL& $0.00$ & $67.65$ & $0.580$ & $0.605$ & $1.185$ & $0.00$ & - & $0.00$ & - \\
		MPUR & $\textbf{0.70}$ & $\textbf{351.38}$ & $\textbf{0.886}$ & $0.896$ & $\textbf{1.781}$ &- &- &- &- \\
		No action & $0.27$ & $225.98$ & $0.665$ & $0.668$ & $1.332$ &- &- &- &- \\
		\midrule
		Human & $1.00$ & $378.04$ & $0.821$ &$0.642$ & $1.463$ & - & - &- &- \\
		\bottomrule
	
	\end{tabular}}
	\label{tab:eval}
\end{table}

\subsection{CARLA Experiments}
The second experiment describes an unprotected left-turning task at a busy intersection in Town04 of the CARLA simulator (Version 0.9.10) \citep{CARLA}, as visualized in Figures \ref{fig:planned_traj} (a) and (b). Six other agents, parameterized with different random driving style, are controlled by CARLA's traffic manager and approach the intersection from three directions. The other agents perform random maneuvers at the intersection. That makes the scenario challenging,  as the prediction of uncertain multi-agent interaction is required to successfully navigate to the goal location. 
For training, we generate a dataset using different sub-optimal behavior policies.
We construct two different test settings. In the first one, CARLA in-distribution (ID), the SDV is spawned at the same locations as during training. In the second setting, CARLA out-of-distribution (OOD), initial configurations are changed to generate novel unseen situations for decision-making. 
We evaluate on $\num{30}$  randomly generated scenarios per test setting, and the results are visualized on the right side of Table \ref{tab:eval}.  Appendix \ref{Experimental Setup} provides additional details about the experimental setup and the data generation. 

Regarding question \textit{Q1}, again UMBRELLA improves the success rate of MBOP by a factor of $\num{2.33} $ (CARLA ID) and $\num{2.15}$ (CARLA OOD) while also navigating the intersection faster. \textit{Our insight is that the superior
performance can be attributed to the use of a stochastic dynamics model}. That underlines the importance of modeling different future outcomes in the context of model-based offline RL for automated driving. Appendix \ref{AdditionalExpResults} provides further visualizations of the model's prediction performance. 1-step IL consistently fails and does not re-accelerate once it stops at the intersection. This behavior is attributed to the \textit{inertia problem} \citep*{Codevilla2019}, a common problem in behavior cloning due to causal confusion \citep{CausalConfusion}. Interestingly, while using the 1-step IL as a prior for rolling out state-trajectories, UMBRELLA and MBOP can overcome this issue, which answers question \textit{Q2}. 

Considering question \textit{Q3}, UMBRELLA achieves a higher success rate and lower MST than its ablation in the CARLA ID experiment. However, in the CARLA OOD experiment, the performance of UMBRELLA drops, whereas the pessimistic variant performs nearly identical. Using UMBRELLA-P, the SDV plans for the worst-case model, i.e. the model, which predicts that all other agents decline the right of way to the SDV. The result is a more conservative and reactive behavior in the intersection causing fewer or no collisions. That is also reflected by the higher MST value.

\subsection{Limitations}
\label{Limitations}
This section approaches question \textit{Q4} and outlines the limitation of current model-based offline planning methods in the context of automated driving.

\textbf{Reward Function Mismatch.} While UMBRELLA, UMBRELLA-P, MBOP, and MPUR learn to perform different challenging maneuvers in dense highway traffic, none of these methods reaches human level performance. To further investigate the cause, let us consider the mean reward $\bar{r}$ in Table \ref{tab:eval}. Notice, that all approaches achieve a higher reward than the human baseline. In the context of offline RL, the task of the agent it to outperform the dataset policy in terms of the reward. We conclude that optimizing the reward function proposed by \cite{MPUR} does not necessarily lead to optimization in terms of driving performance. That is caused by the fact that the reward function does not exactly represent the human driving style. As other agents do not react to the SDV, the most likely way to get a success rate of $1.00$ is to drive exactly like the human in the dataset. One remedy, is to learn the reward function with inverse RL (IRL) methods \citep{inverseRL}. However, this raises the almost philosophical question whether a human-like reward function is the ultimate goal for automated driving, as humans sometimes also tend to drive in a safety-critical way and violate traffic rules. In general, reward (mis)design is a common problem in automated driving contributions \citep{Rewardmisdesign} and needs to be addressed in future work. We hypothesize that this effect also occurs in our CARLA experiments.

\textbf{Dependency on BC Policy.}
While UMBRELLA and MBOP can improve upon a simple BC policy, we observed that their performance is limited by the unimodal BC policy. That often leads to sampled trajectories, which correspond to one possible maneuver (see Figure \ref{fig:planned_traj} (d)), whereas in reality a human driver plans, for multiple possible maneuvers. Therefore, future work should investigate the integration of multi-modal BC policies.   
In addition, the CARLA dataset contains suboptimal expert policies, which drive through the intersection aggressively and are mostly non-reactive. We observed that this driving style is partially also present in MBOP's and UMBRELLA's policies, which shows a downside of both approaches when learning from non-expert-like datasets.

\section{Conclusion}
This work proposed an approach for model-based offline reinforcement learning, which considers aleatoric and epistemic uncertainty. Experiments in multiple challenging automated driving scenarios with dense traffic showed that incorporating aleatoric uncertainty improves the performance of planning-based approaches. Moreover, the proposed method improves upon the poor performance of simple BC policies. A pessimistic ablation is inferior in scenarios of the training distribution but seems to offer advantages in out-of-distribution scenarios.
We also showed the limitations of the proposed approach and current state-of-the-art planning-based offline RL in the context of automated driving. Future work should focus on improving the BC policy priors, use graph-based representations instead of images and investigate the use of other optimization techniques.

\begin{ack}
	This research was funded by the Federal Ministry for Economic Affairs and Energy on the basis of a decision by the German Bundestag in the project “KISSaF – AI-based Situation Interpretation for Automated Driving”.
\end{ack}

\bibliographystyle{plainnat}
\bibliography{neurips_2021_ml4ad_references}


\newpage
\appendix

\section{Model Architecture}
\label{Model Architecture}

\subsection{Dynamics Model}

The dynamics model is inspired by the work of \cite{MPUR} and its architecture is visualized in Figure \ref{fig:fm_arch}. The network consists of the three main components: the encoder $f_{\mathrm{enc}}$, the posterior network $q_{\boldsymbol{\phi}}$, and the decoder $f_{\mathrm{dec}}$. For reasons of clarity, we do not include latent dropout (see Appendix \ref{TdetailsDynamics})  in the figure. The illustration is further simplified as in reality, the posterior network outputs the parameters of a Gaussian distribution, which are then used to estimate $\textbf{z}_t$ by 
\begin{equation}
	\label{eq:update_fm1}
	(\boldsymbol{\mu}_{\boldsymbol{\phi}}, \boldsymbol{\sigma}_{\boldsymbol{\phi}}) = q_{\boldsymbol{\phi}}(\textbf{s}_t, \textbf{s}_{t+1})
\end{equation}
\begin{equation}
	\label{eq:update_fm2}
	\boldsymbol{\epsilon} \sim \mathcal{N}(0,I)
\end{equation}
\begin{equation}
	\label{eq:update_fm3}
	\textbf{z}_t = \boldsymbol{\mu}_{\boldsymbol{\phi}} + \boldsymbol{\sigma}_{\boldsymbol{\phi}} * \boldsymbol{\epsilon}.
\end{equation}The work of \cite{MPUR} gives further information about the layer sizes and different parts of the model.

\subsection{BC Policy and Truncated Value Function}
\begin{figure}[p]
	\centering
	\resizebox{0.8\textwidth}{!}{
		\begin{tikzpicture}
			\node[anchor=south west,inner sep=0] (image) at (0,0) {\includegraphics[width=0.85\textwidth]{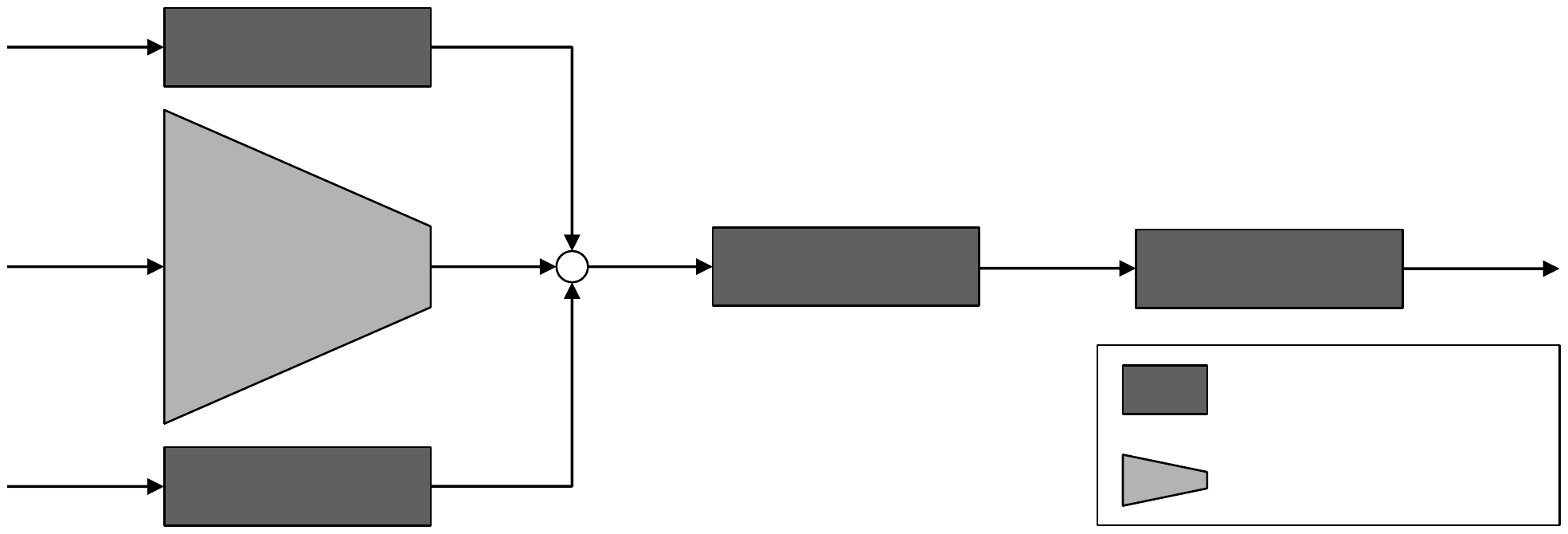}};
			\node[anchor=west](mlp) at (9.2,1.05) {2-layer MLP};
			\node[anchor=west](conv) at (9.2,0.35) {3-layer ConvNet};
			
			\node[anchor=east](pastAc) at (0.1,3.8) {$\textbf{a}_{(t-n_c):(t-1)}$};
			\node[anchor=east](pastIm) at (0.1,2.1) {$\textbf{i}_{(t-n_c+1):t}$};
			\node[anchor=east](pastSt) at (0.1,0.35) {$\textbf{u}_{(t-n_c+1):t}$};
			\node[anchor=east](action) at (12.5,2) {$\hat{R}_H$};
		\end{tikzpicture}
	}
	\caption{Truncated value function network architecture.}
	\label{fig:vf_arch}
\end{figure}

The network architecture of the BC policy and the Truncated Value Function are visualized in Figure \ref{fig:bc_arch} and \ref{fig:vf_arch}. They encode the sequence of images with a 3-layer convolutional neural network with (64 -128- 256) feature maps. The sequences of measurement vectors $\textbf{u}_t$ and actions $\textbf{a}_t$ are encoded using a 2-layer fully connected network with $\num{256}$ units. The addition of all encoding is then further processed by two 2-layer MLPs. The result is the predicted action $\hat{\textbf{a}}_t$ or return $\hat{\textbf{R}}_t$, respectively.
\FloatBarrier
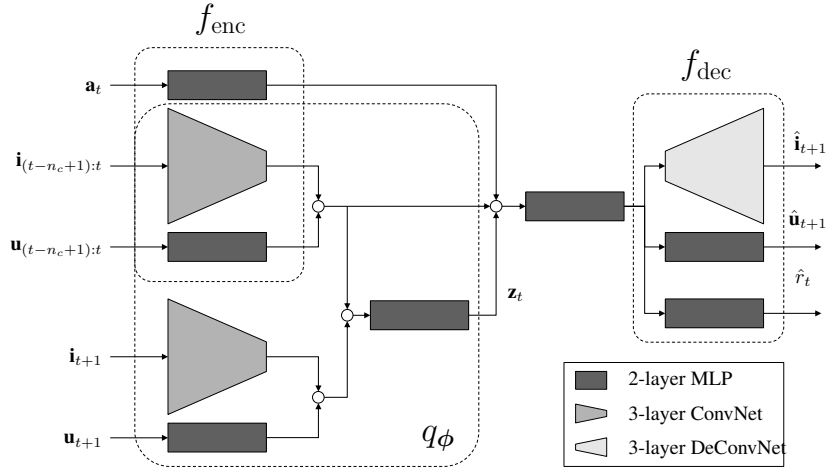
\begin{figure}[p]
	\centering
	\resizebox{0.8\textwidth}{!}{
		\begin{tikzpicture}
			\node[anchor=south west,inner sep=0] (image) at (0,0) {\includegraphics[width=0.9\textwidth]{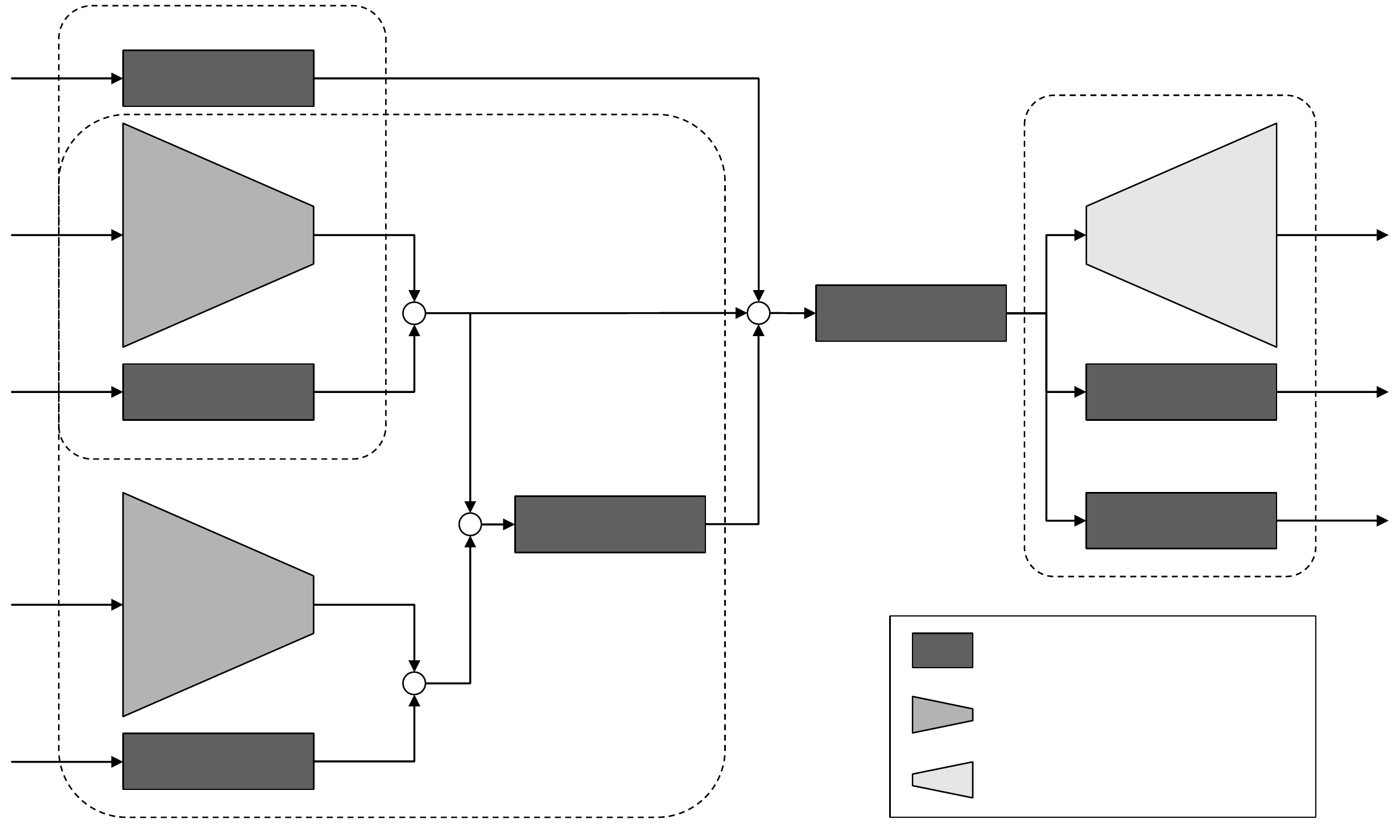}};
			\node[anchor=west](mlp) at (9,1.7) {2-layer MLP};
			\node[anchor=west](conv) at (9,1.1) {3-layer ConvNet};
			\node[anchor=west](deconv) at (9,0.5) {3-layer DeConvNet};
			
			\node[anchor=west](fenc) at (1.45,8) {\LARGE $f_{\mathrm{enc}}$};
			\node[anchor=west](fdec) at (9.9,7.2) {\LARGE $f_{\mathrm{dec}}$};
			\node[anchor=west](posterior) at (5.4,0.7) {\LARGE $q_{\boldsymbol{\phi}}$};
			
			\node[anchor=east](action) at (0.1,6.8) {$\textbf{a}_t$};
			\node[anchor=east](pastIm) at (0.1,5.5) {$\textbf{i}_{(t-n_c+1):t}$};
			\node[anchor=east](pastSt) at (0.1,4) {$\textbf{u}_{(t-n_c+1):t}$};
			
			\node[anchor=east](trueIm) at (0.1,2.1) {$\textbf{i}_{t+1}$};
			\node[anchor=east](trueSt) at (0.1,0.65) {$\textbf{u}_{t+1}$};
			
			\node[anchor=east](predIm) at (0.91\textwidth,5.85) {$\hat{\textbf{i}}_{t+1}$};
			\node[anchor=east](predSt) at (0.91\textwidth,4.5) {$\hat{\textbf{u}}_{t+1}$};
			\node[anchor=east](predRe) at (0.89\textwidth,3.5) {$\hat{r}_{t}$};
			
			\node[anchor=west](z) at (6.9,3.2) {$\textbf{z}_{t}$};	
		\end{tikzpicture}
	}
	\caption{Simplified network architecture of the stochastic forward dynamics model during training}
	\label{fig:fm_arch}
\end{figure}
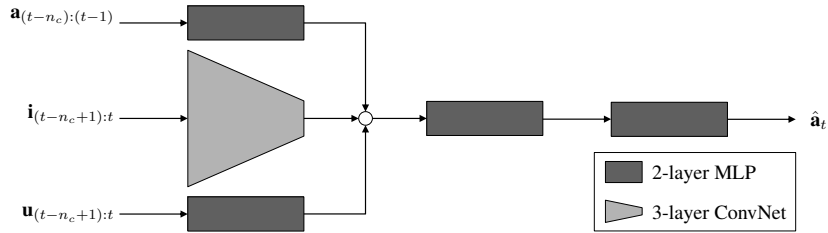
\begin{figure}[p]
	\centering
	\resizebox{0.8\textwidth}{!}{
		\begin{tikzpicture}	
			\node[anchor=south west,inner sep=0] (image) at (0,0) {\includegraphics[width=0.85\textwidth]{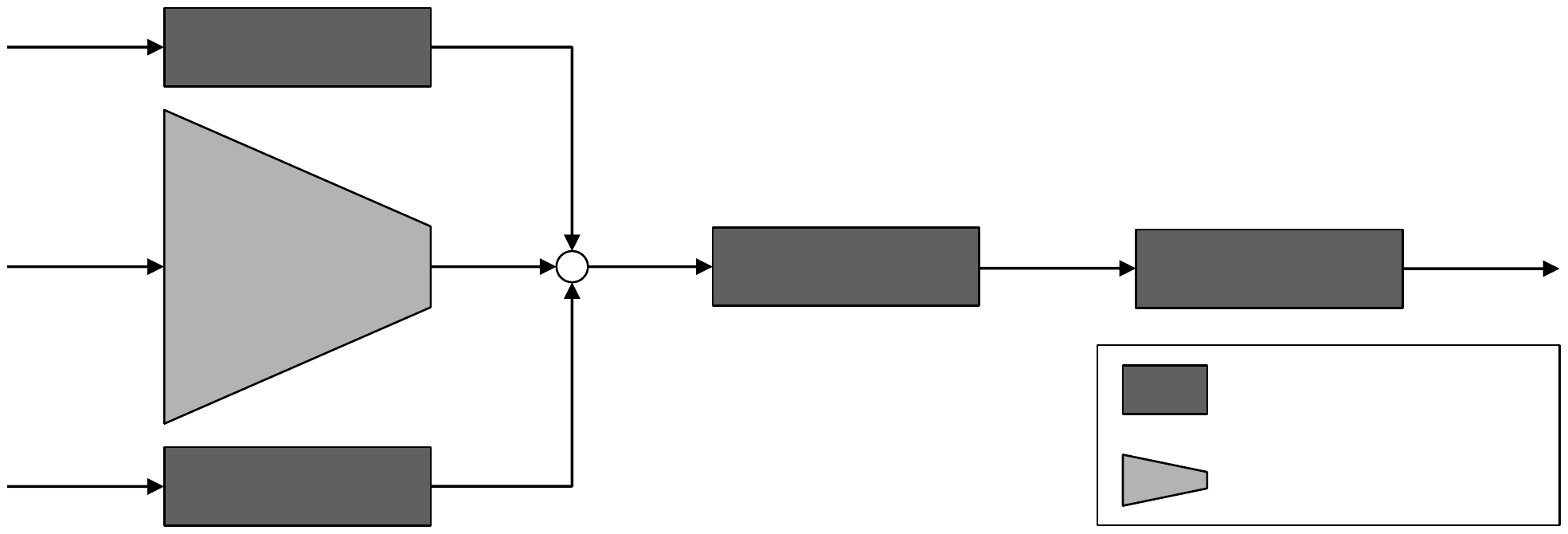}};
			\node[anchor=west](mlp) at (9.2,1.05) {2-layer MLP};
			\node[anchor=west](conv) at (9.2,0.35) {3-layer ConvNet};
			
			\node[anchor=east](pastAc) at (0.1,3.8) {$\textbf{a}_{(t-n_c):(t-1)}$};
			\node[anchor=east](pastIm) at (0.1,2.1) {$\textbf{i}_{(t-n_c+1):t}$};
			\node[anchor=east](pastSt) at (0.1,0.35) {$\textbf{u}_{(t-n_c+1):t}$};
			\node[anchor=east](action) at (12.5,2) {$\hat{\textbf{a}}_t$};
		\end{tikzpicture}
	}
	\caption{Behavior-cloned policy network architecture.}
	\label{fig:bc_arch}
\end{figure}
\begin{figure}[p]
	\centering
	\resizebox{0.7\textwidth}{!}{
		\begin{tikzpicture}
			\node[fill=white, draw=black,circle] (st) {$\textbf{s}_t$};
			\node [right=of st, minimum width=1cm, minimum height=0.6cm, fill=white, draw=black, rectangle, rounded corners=4pt,  align=center] (enc) {$f_{\mathrm{enc}}$};
			\node[below of=enc, fill=white, draw=black,circle] (at) {$\textbf{a}_t$};
			\node[right=of enc, fill=white, draw=black, circle, minimum width=0.4cm, minimum height=0.4cm] (sum){};
			\node [right=2cm of sum, minimum width=1cm, minimum height=0.6cm, fill=white, draw=black, rectangle, rounded corners=4pt,  align=center] (dec) {$f_{\mathrm{dec}}$};
			\node[right=2cm of dec, fill=white, draw=black,circle, scale=0.9] (st1) {$\hat{\textbf{s}}_{t+1}$};
			\node[below of=st1, fill=white, draw=black,circle,scale=0.9] (rt1) {$\hat{\textbf{r}}_{t+1}$};
			\node[right= of rt1, minimum width=1cm, minimum height=0.6cm, fill=white, draw=black, rectangle, rounded corners=4pt,  align=center, dashed] (mseR) {MSE};
			\node[above=of sum, fill=white, draw=black,circle,  yshift=-0.6cm] (zt) {$\textbf{z}_{t}$};
			\node [above=of st1, minimum width=1cm, minimum height=0.6cm, fill=white, draw=black, rectangle, rounded corners=4pt,  align=center, dashed,  yshift=0.3cm] (mse) {MSE};
			\node[above=of mse, fill=white, draw=black,circle,  yshift=0.3cm, scale=0.9] (realst1) {$\textbf{s}_{t+1}$};
			\node[right= of realst1, above= of mseR, fill=white, draw=black,circle,scale=0.9] (realrt1) {$\textbf{r}_{t+1}$};
			
			\node[left=2cm of  realst1, minimum width=1cm, minimum height=0.6cm, fill=white, draw=black, rectangle, rounded corners=4pt,  align=center] (q) { $q_{\boldsymbol{\phi}}(\textbf{z} \mid \textbf{s}_t, \textbf{s}_{t+1})$};
			\node [below=of q , minimum width=1cm, minimum height=0.6cm, fill=white, draw=black, rectangle, rounded corners=4pt,  align=center, dashed,  yshift=0.55cm] (kl) {$D_{\mathrm{KL}}$};
			\node[below=0.5cm of kl, minimum width=1cm, minimum height=0.6cm, fill=white, draw=black, rectangle, rounded corners=4pt,  align=center,  yshift=0.1cm] (p) {$p(\textbf{z})$};
			\node[above=of zt, fill=white, draw=black,circle, minimum size=1pt, yshift=-0.6cm, scale=0.3] (src_switch) {};
			\node[left=of p, fill=white, draw=black,circle, minimum size=1pt, scale=0.3] (end_switch_p) {};
			\node[above=of src_switch, fill=white, draw=black,circle, minimum size=1pt, yshift=-0.7cm, xshift=-0.38cm, scale=0.3] (end_switch_q) {};
			
			\draw[->] (st) -- (enc); 
			\draw[->] (at) -- (enc);
			\draw[->] (enc) -- node[below, near end, font=\tiny] {$+$} (sum);
			\draw[->] (sum) -- (dec);
			\draw[->] (dec) -- (st1);
			\draw[->] (dec.south) |- (rt1);
			\draw[->,white!82.7450980392157!black] (q) -| node[left, above, near start, xshift=-0.2cm, font=\tiny, black]{Sample} (end_switch_q);
			\draw[->] (p) -- node[left, above, near start, xshift=-0.2cm,  font=\tiny]{Sample} (end_switch_p);
			\draw[-,white!82.7450980392157!black] (end_switch_q) -- node[left, xshift=-0.2cm, font=\tiny, text width=0.7cm, yshift=-0.1,black] {Latent dropout} (src_switch);
			\draw[-] (src_switch) -- (end_switch_p);
			\draw[->] (src_switch) -- (zt);
			\draw[->] (zt) -- node[left, near end, font=\tiny] {$+$} (sum);
			\draw[->,white!82.7450980392157!black] (st1) -- (mse);
			\draw[->,white!82.7450980392157!black] (realst1) -- (mse);
			\draw[->,white!82.7450980392157!black] (realst1) -- (q);
			\draw[->,white!82.7450980392157!black] (realrt1) -- (mseR);
			\draw[->,white!82.7450980392157!black] (rt1) -- (mseR);
			\draw[->,white!82.7450980392157!black] (p) -- (kl);
			\draw[->,white!82.7450980392157!black] (q)  -- (kl);
			\draw[->,white!82.7450980392157!black] (st) -- ++(0,5) -| (q);
		\end{tikzpicture}
	}
	\caption{Signal flow of the stochastic forward dynamics model (inspired by \cite{MPUR}). Black color denotes the signal flow during training and inference. Grey denotes the signal flow, which is only used during training.}
	\label{fig:vae}
\end{figure}
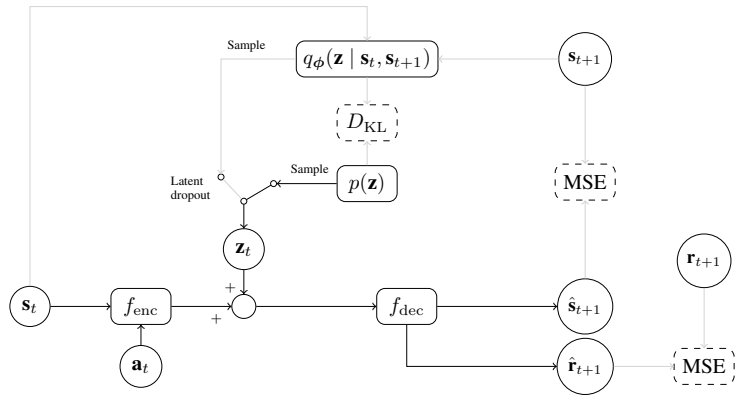

\section{Training Details}
\label{Training Details}
\subsection{Dynamics Model}
\label{TdetailsDynamics}
 The signal flow during training and testing is visualized in Figure \ref{fig:vae}. The network $f_{\mathrm{enc}}$ encodes the current state $\textbf{s}_t$ and action $\textbf{a}_t$. Moreover, a latent variable $\textbf{z}_t$ is added. $f_{\mathrm{dec}}$ decodes the result leading to the prediction of the successor state $\hat{\textbf{s}}_{t+1}$ and reward $\hat{r}_{t+1}$. During training, the mean squared error (MSE) between the predicted successor state $\hat{\textbf{s}}_{t+1}$ and the true successor state $\textbf{s}_{t+1}$ is minimized. The same holds for the reward. The network further encodes the current state $\textbf{s}_t$ and the true successor state $\textbf{s}_{t+1}$ with $f_{\mathrm{enc}}$ to the parameterized posterior distribution  $q_{\boldsymbol{\phi}}(\textbf{z}\mid \textbf{s}_t, \textbf{s}_{t+1})$. Besides the MSE losses, the Kullback-Leibler divergence ($D_{\mathrm{KL}}$) between the posterior and the prior distribution $p(\textbf{z})$ is minimized during training.
 To make the network sensitive to the action input, we apply \textit{latent dropout} \citep{MPUR} during training. First, a random variable $x$ is sampled from a Bernoulli distribution $\mathcal{B}(p_x)$ with probability $p_{x}$. Hence we sample the latent variable during training according to \begin{equation}
	\textbf{z}_t \sim (1-x)*q_{\boldsymbol{\phi}}(\textbf{z} \mid \textbf{s}_t, \textbf{s}_{t+1}) + x*p(\textbf{z})
\end{equation}
That forces the model to receive as much information as possible from the current state and action to predict the successor state. To mitigate aggregated multi-step prediction errors \citep{moerland2021modelbased} at test time, we unroll the model for five timesteps and train using the corresponding loss.
The dynamics model is first trained in deterministic mode for $\num{50000}$ steps. Deterministic mode denotes the training configuration, where no latent variable is added. Afterward, the dynamics model is again trained for $\num{50000}$ steps with the added latent variable. We denote this model as the {stochastic dynamics model}. The \textit{determinstic dynamics model} used for MBOP is trained without the latent variable addition. 

\subsection{BC Policy and Truncated Value Function}
We train the BC policy network by minimizing the loss \begin{equation}
	\label{eq:loss_bc}
	\mathcal{L}(\boldsymbol{\psi}; \textbf{s}_t,\textbf{a}_{(t-n_c):(t-1)}, \textbf{a}_t) = {|| \textbf{a}_{t} - f_{\mathrm{b},\boldsymbol{\psi}}(\textbf{s}_t,\textbf{a}_{(t-n_c):(t-1)})||}_2^2, 
\end{equation}
whereas parameter vector of one behavior-cloned policy model is denoted by $\boldsymbol{\psi}$.
The parameterized truncated value function model  $f_{\mathrm{R},\boldsymbol{\xi}}(\textbf{s}_t,\textbf{a}_{(t-n_c):(t-1)})$ is learned by minimizing the following per-sample loss:
\begin{equation}
	\label{eq:loss_vf}
	\mathcal{L}(\boldsymbol{\xi}; R_H, \textbf{s}_t,\textbf{a}_{(t-n_c):(t-1)}) = {|| R_H - f_{\mathrm{R},\boldsymbol{\xi}}(\textbf{s}_t,\textbf{a}_{(t-n_c):(t-1)})||}_2^2 
\end{equation}
 $\boldsymbol{\xi}$ is the parameter vector. The value function target $R_H\in\mathbb{R}$ is defined as the sum over the next $H$ rewards:
\begin{equation}
	R_H = \sum_{t=\tau}^{t+H-1}{r_t}
\end{equation}

Table \ref{tab:training_param} lists the hyperparameters for training the stochastic dynamics model, the BC policy model, and the value function model.

\begin{table}[th]
	\centering
	\caption{Training hyperparameters used for the stochastic forward dynamics model, the BC policy model, and the value function model. $n_p$ denotes the number of time steps the model is unrolled during training. $\eta$ is the initial learning rate. $p_\textrm{dropout}$ describes the dropout probability. $p_{x}$ denotes the latent dropout probability}
	\label{tab:training_param}
	\resizebox{\textwidth}{!}{
		\begin{tabular}{l c c c c c c c c c}
			\toprule
			\\[-1em]
			Model & Type & $n_c$ & $n_p$ & Opt. & Steps & Batchsize & $\eta$ & $p_\textrm{dropout}$ & $p_{x}$ \\
			\midrule
			Dynamics & stochastic & $20$ & $5$ & Adam & $2*50000$ & $32$ & $0.0001$ & $0.1$ & $0.5$ \\
			BC Policy & deterministic & $20$ & $1$ & Adam & $50000$ & $32$ & $0.0001$ & $0.0$ & - \\
			Value Func. & deterministic & $20$ & $30$ & Adam & $50000$ & $32$ & $0.0001$ & $0.1$ & - \\
			\bottomrule
		\end{tabular}
	}
\end{table}

\section{Experimental Setup}
\label{Experimental Setup}

\subsection{NGSIM}
\textbf{State Space}.
The state $\textbf{s}_t$ at time $t$ is approximated by a set of $n_\textrm{c}$ observations $\textbf{o}_t$. One observation consists of a context image in bird's-eye-view $\textbf{i}_t\in \mathbb{R}^{4\times H \times W}$ and a measurement vector $\textbf{u}_t\in \mathbb{R}^4$. The context image has sizes $H=117$ pixel and $W=24$ pixel and is centered around the SDV. It contains four channels with information about the SDV's size, the drivable area, lane markings, and the surrounding vehicles. The resolution is approximately $0.5$ meter per pixel. The measurement vector contains information about the 2D position $\textbf{p}_t \in \mathbb{R}^2$ and velocity $\textbf{v}_t \in \mathbb{R}^2$ of the SDV.

\textbf{Action Space.}
The action $\textbf{a}_t = (\Delta v_t, \Delta \delta_t)$ describes a change of velocity $\Delta v_t \in \mathbb{R}$ and a change of steering angle $\Delta \delta_t \in \mathbb{R}$ by 
\begin{equation}
	\Delta v_t = ||\textbf{v}_{t+1}||_2 - ||\textbf{v}_{t}||_2
\end{equation}
\begin{equation}
	\Delta \delta_t = (\textbf{v}_{t+1} - \textbf{v}_{t})^T \frac{(\textbf{v}_t)_\perp}{||\textbf{v}_{t}||_2}
\end{equation}

\textbf{Reward.}
The SDV optimizes the following reward function, which was proposed by \cite{MPUR}:
\begin{equation}
	r_t = r_{\textrm{prox},t} + r_{\textrm{lane},t}
\end{equation}  
 It penalizes small distances to other objects scaled by the SDV's velocity (proximity reward $ r_{\textrm{prox},t}$) and small distances to the lane markings (lane reward $r_{\textrm{lane},t}$).

 \textbf{Dataset}. A total of 5596 vehicle trajectories sampled at 10Hz are included. Note that one trajectory corresponds to one episode. We split the trajectories into training ($\num{80}\%$), validation ($\num{10}\%$) and testing sets ($10\%$). 
 
\subsection{CARLA}

\textbf{State Space.}
The state space is the same as in the NGSIM experiment besides that $H=126$, $W=74$ and $\textbf{i}_t\in \mathbb{R}^{3\times H \times W}$. In this experiment, we excluded the ego-layer as the SDV always has the same size during training. Note that this is not the case in the NGSIM experiment due to different vehicle types and sizes.

\textbf{Action Space.}
The action $\textbf{a}_t = (d_t, \delta_t)$ describes a normalized lateral steering wheel command $\delta_t \in[-1, 1]$ and longitudinal driving command $d_t \in[-1, 1]$ (acceleration or braking), which is converted to a throttle and brake command using a low-level PID controller.

\textbf{Reward.}
The SDV optimizes the following reward function: 
\begin{equation}
	r_t = w_1*r_{\textrm{prog},t} + w_2*r_{\textrm{lane},t} + w_3*r_{\textrm{coll},t}.
\end{equation} $r_{\textrm{prog},t}$ rewards the SDV's progress towards its goal while obeying the speed limit. The lane reward $r_{\textrm{lane},t}$ penalizes deviations from the reference lane center. $r_{\textrm{coll},t}$ is a collision reward encouraging the agent to avoid collisions. Note that $w_1, w_2, w_3\in \mathbb{R}$ are weighting factors. Further, the three reward terms are defined as:
\newcommand\numberthis{\addtocounter{equation}{1}\tag{\theequation}}
\begin{align*} 
	r_{\textrm{prog},t}&=
	\begin{cases}
		\gamma^t*(e_t-e_{t-1}), & \text{if }  ||\textbf{v}_t||_2^2 \leq v_{\textrm{limit}}\\
		0, & \text{otherwise}
	\end{cases} \numberthis \label{eq:r_progress} \\ 
	r_{\textrm{lane},t}&=-||\textbf{p}_t-\textbf{p}_{\textrm{center}}||_2^2 \numberthis \label{eq:r_lane}\\
	r_{\textrm{coll},t}&=
	\begin{cases}
		-2+0.2(t_{\textrm{coll}}-t), & \text{if } t \in \left[t_{\textrm{coll}}-9, t_{\textrm{coll}}\right]\\
		0, & \text{otherwise}
	\end{cases} \numberthis \label{eq:r_collision}
\end{align*}
$e_t\in \mathbb{R}$ it the distance to the target location and $\gamma = 0.99$ a discount factor.  $v_{\textrm{limit}}\in\mathbb{R}$ describes the speed limit. The lane reward is defined as the negative Euclidean distance to the lane center defines the lane reward. The global position of the lane center closest to the position of the SDV is given by $\textbf{p}_{\textrm{center}}$. If a collision occurs in time step $t$ the collision reward also penalizes the prior $9$ time steps according to a linear function. $t_{\textrm{coll}}\in\mathbb{R}$ denotes the collision timestamp. That should make the agent learn dangerous pre-collision states. This work chooses $w_1=1, w_2=0.1, w_3=1$.

\textbf{Dataset.} The models are trained based on trajectories corresponding to $\num{1200}$ episodes, which are recorded by driving the intersection with different policies. We intentionally make the policies sub-optimal to see how the different algorithms can handle non-expert-like data. This work uses a total of three different behavior policies because drivers in real-world datasets also follow different driving styles. The first policy controls the SDV using PID-controllers, which track the path given by a route planning module. All agents are spawned according to the parameters in Table \ref{tab:data_gen_params}. To generate a diverse set of scenarios the parameters are randomly drawn. Other agents are controlled using the autopilot of the CARLA traffic manager. At the intersection they, follow a First-In-First-Out (FIFO) order. The SDV's policy can be randomized by three different driving styles: cautions, normal and aggressive. This behavior affects the maximum speed and maintained safety distance to other agents on its planned path. The other agent's behavior is further randomized by changing their traffic sign, other vehicles, and speed limit ignorance rate.  When the SDV reaches the intersection, it waits for a random amount of time $t_\textrm{wait} \in \left[0\si{\second},15\si{\second}\right]$ before starting the turn. Note, that this often causes collisions, as the SDV does not react to vehicles that approach from the side. Hence, we filter out collision-prone episodes during data generation. However, the recorded driving behavior is still sub-optimal as the SDV assumes it always has the right of way, which results in a safety-critical behavior. The second policy drives like the first one, but random noise is applied to the controls while maneuvering the intersection. For this purpose, a random offset is injected for $\num{5}$ time steps. This work uses both policies to generate $\num{400}$ collision-free episodes each. Lastly, another $\num{400}$ collision-free episodes are constructed by a human driver, who controls the car with the keyboard. Due to difficult control the policy is also sub-optimal. 
\begin{table}[th]
	\centering
	\caption{Randomized parameters for data generation.}
	\begin{tabular}{l c c}
		\toprule
		&SDV & Other agents \\
		\midrule
		\thead{Spawn position $[\si{\metre}]$} 
		& \thead{$[25,45]$} & \thead{first vehicle of each direction: $[25,35]$, \\ second vehicle of each direction: $[45,50]$} \\
		\thead{Agent behavior} & \thead{[cautious, normal, aggressive]} & \thead{safety distance $[\si{\metre}]$: $[0.1, 10]$ \\ sign ignorance: $[0, 0.2]$, \\ vehicle ignorance: $[0, 0.2]$\\  speed difference: $[-0.3, 0.3]$} \\
		\thead{Waiting time at intersection $[\si{\second}]$} & \thead{$[0,15]$} & \thead{FIFO} \\
		\thead{Noise injection offset} & \thead{throttle: $[-0.5,0.5]$\\ brake: $[-0.5,0.5]$\\ steering: $[-0.05,0.05]$}& \thead{-}\\
		
		\bottomrule
	\end{tabular}
	\label{tab:data_gen_params}
\end{table}

\subsection{Planning Hyperparameters}

We use the hyperparameter of Table \ref{tab:planning_param} during our experiments, which were found using a grid search. Note that, Off Policy Evaluation \citep{OffPolicyEvaluation} and Offline Hyperparameter Selection \citep{OfflineHPS} are outside the scope of this work. For a fair comparison we performed a search for all used models during evaluation.

\begin{table}[th]
	\centering
	\caption{Used hyperparameters of UMBRELLA for the NGSIM and CARLA experiments.}
	\label{tab:planning_param}
		\begin{tabular}{l c c c c c c }
			\toprule
			Environment & $K$ & $H$ & $N$ & $\sigma^2$ & $\beta$ & $\kappa$  \\
			\midrule
			NGSIM & $2$ & $30$ & $300$ & $1.2$ & $0.6$ & $0.5$ \\
			CARLA & $2$ & $30$ & $100$ & $1.5$ & $0.9$ & $0.5$ \\
			\bottomrule
		\end{tabular}
	
\end{table}

\section{Additional Experimental Results}
\label{AdditionalExpResults}

\subsection{Video Prediction Results}
\begin{figure}[!t]
	\centering
	\begin{subfigure}{\linewidth}
		\centering
		\includegraphics[width=0.1\linewidth]{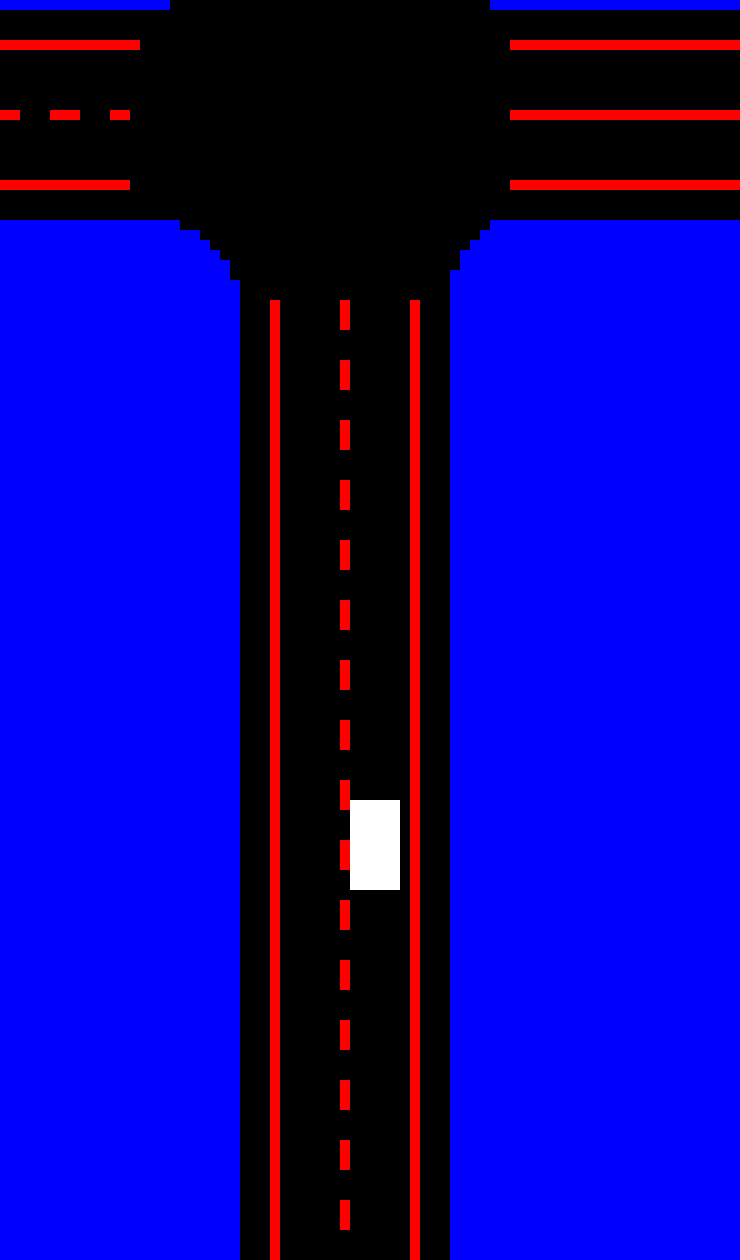}\hfill
		\includegraphics[width=0.1\linewidth]{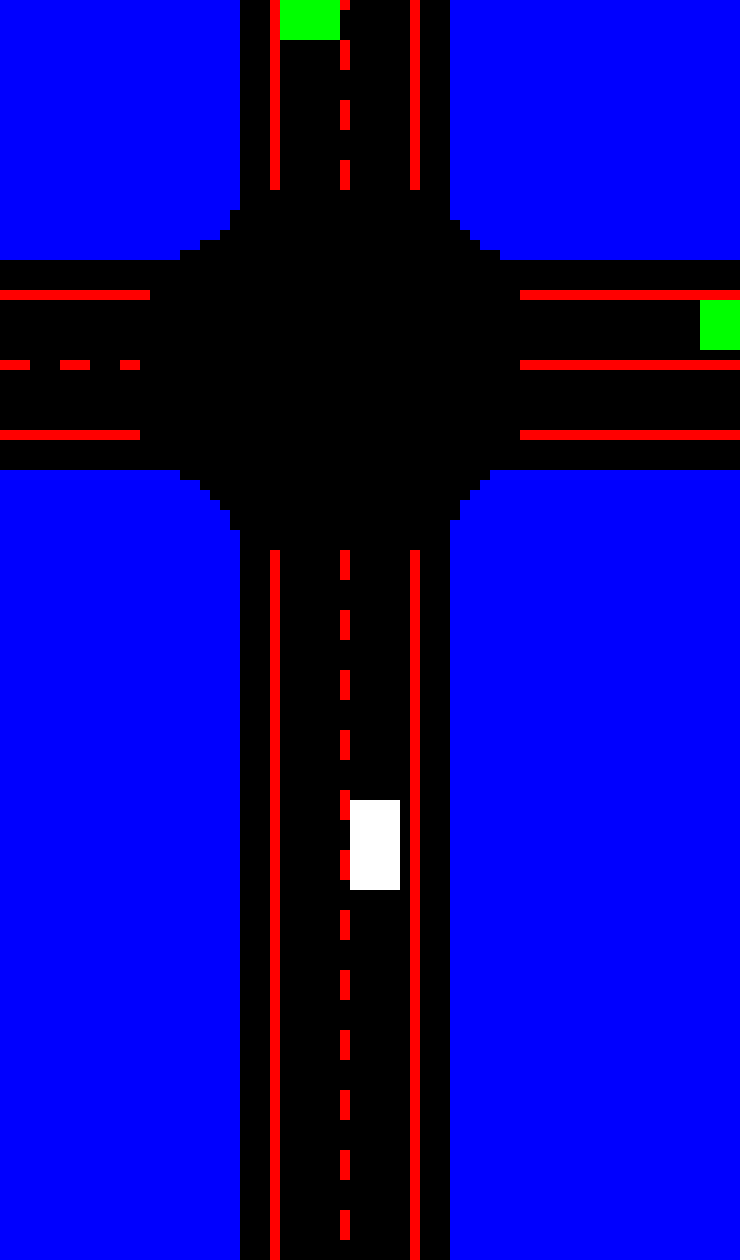}\hfill
		\includegraphics[width=0.1\linewidth]{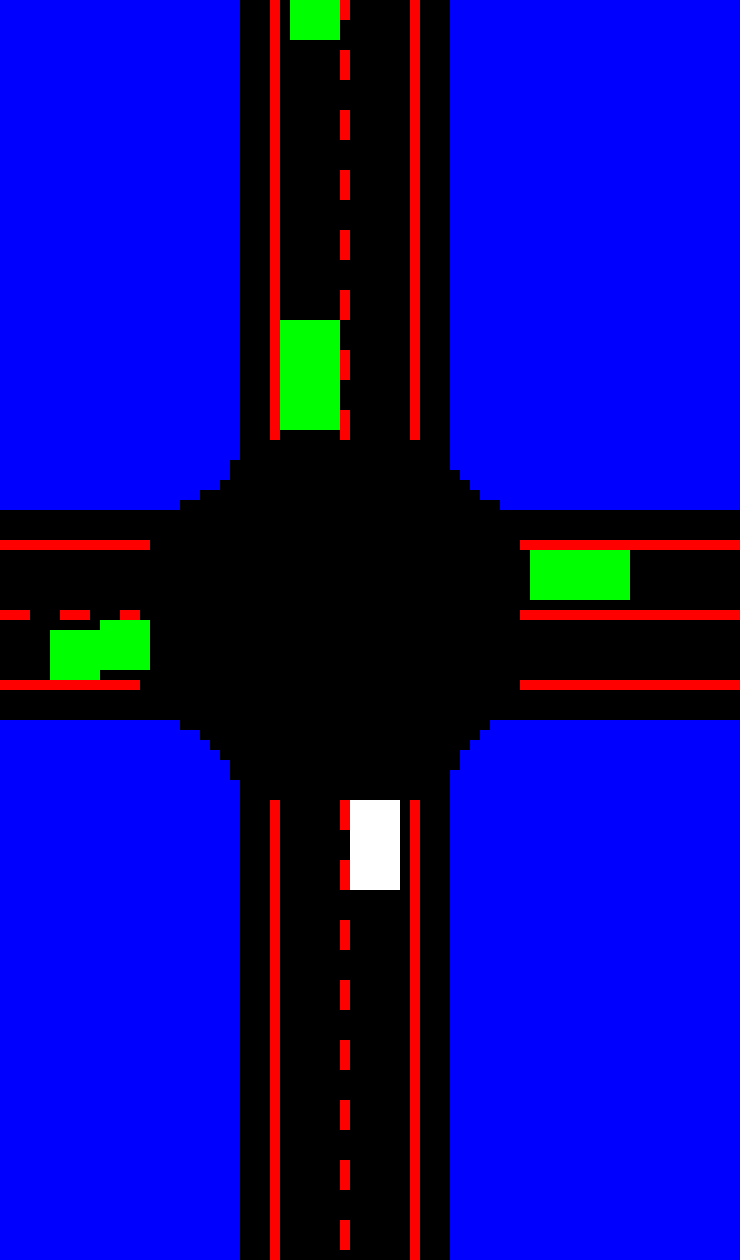}\hfill
		\includegraphics[width=0.1\linewidth]{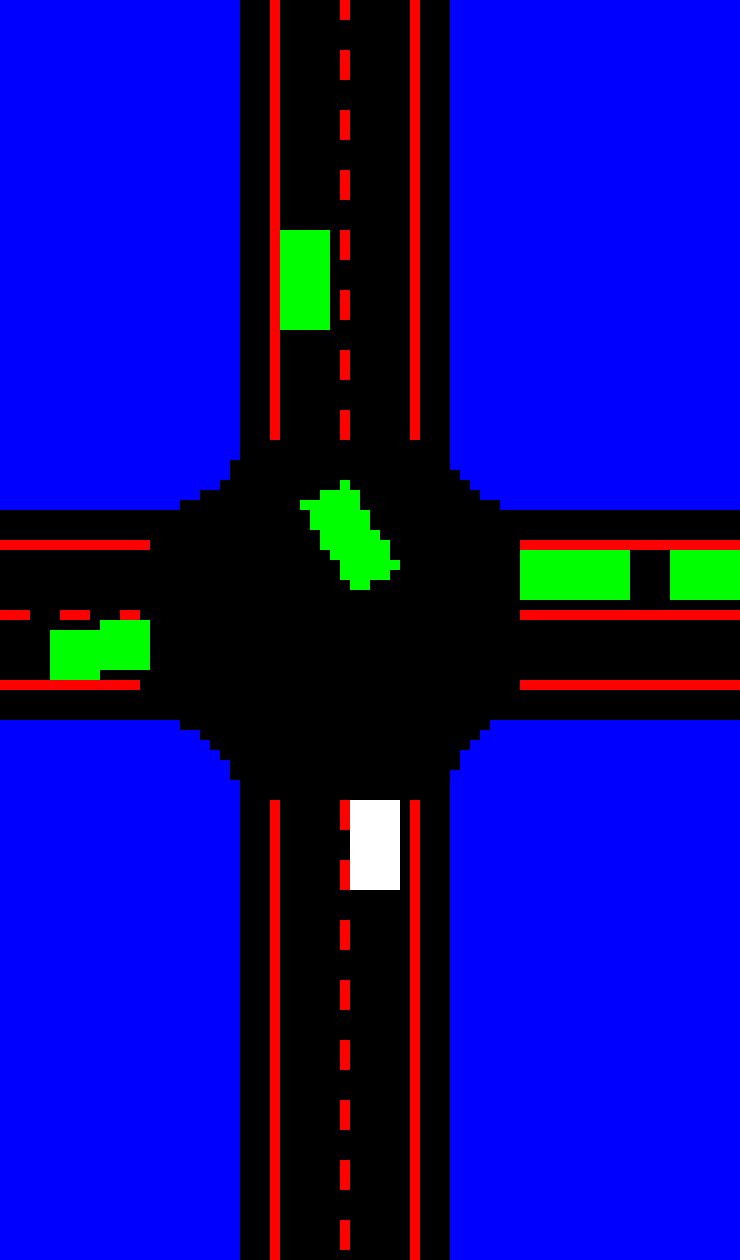}\hfill
		\includegraphics[width=0.1\linewidth]{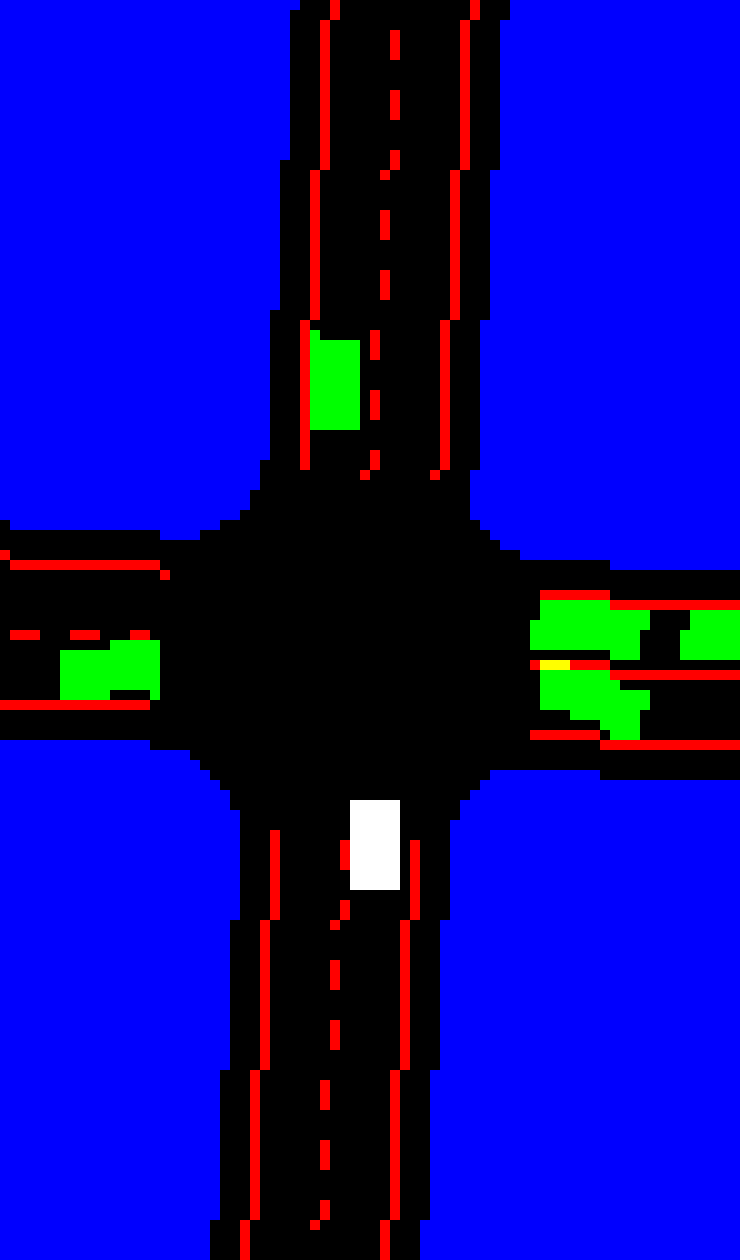}\hfill
		\includegraphics[width=0.1\linewidth]{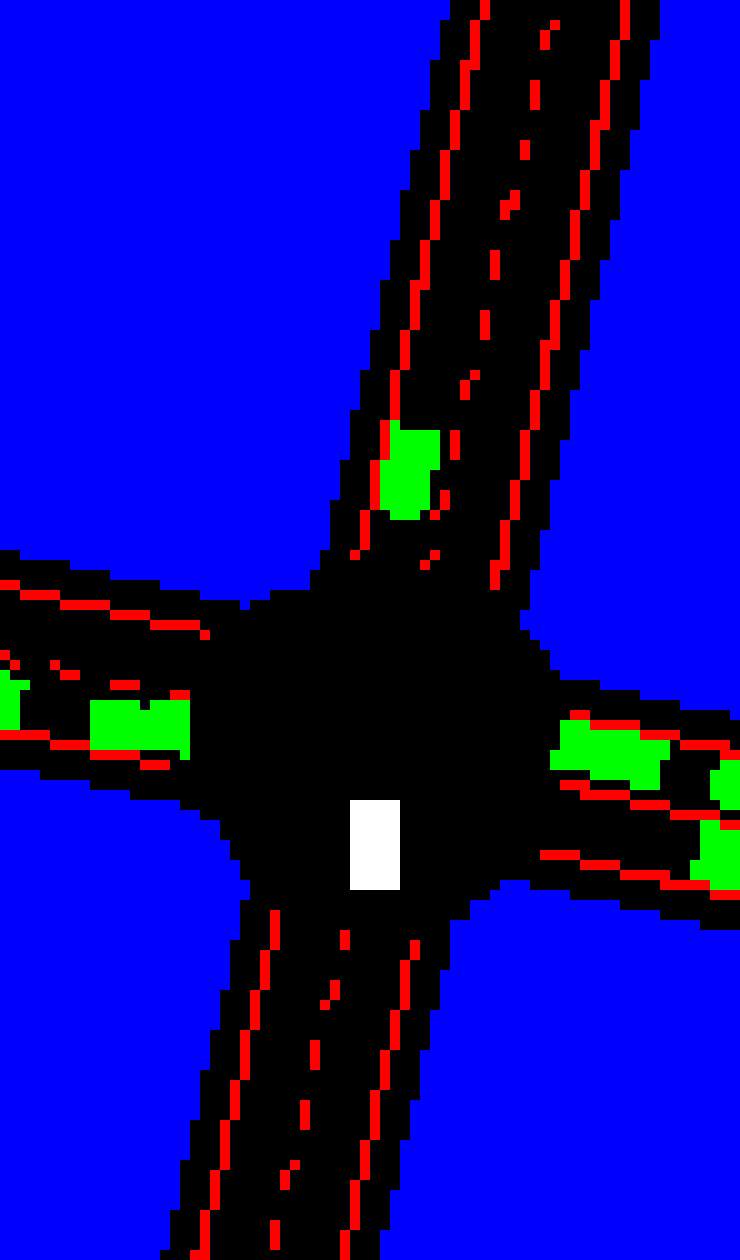}\hfill
		\includegraphics[width=0.1\linewidth]{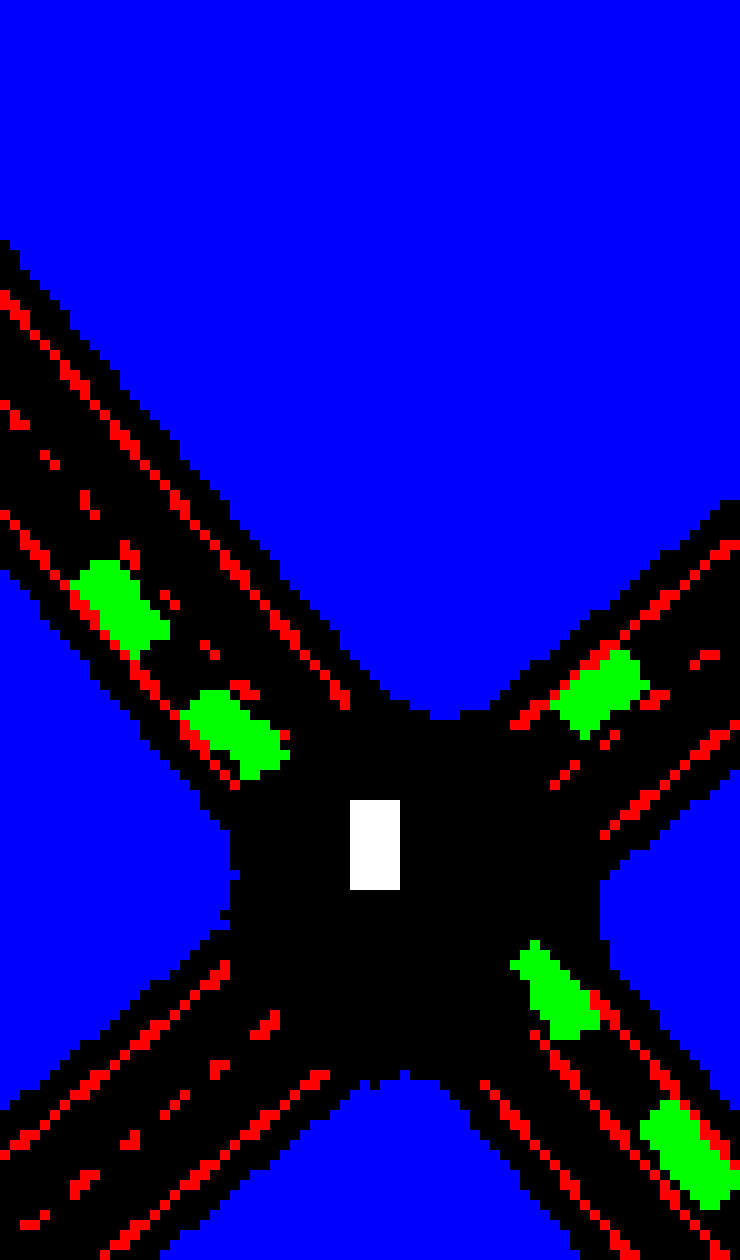}\hfill
		\includegraphics[width=0.1\linewidth]{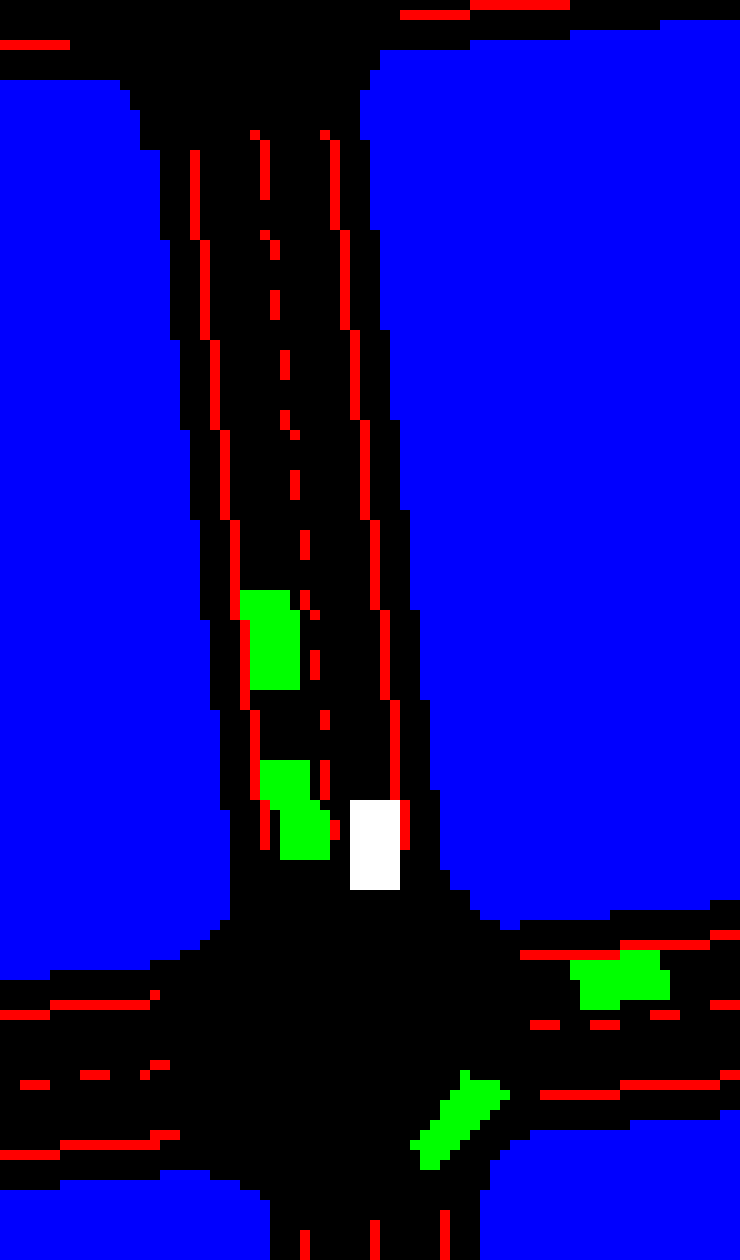}\hfill
		\includegraphics[width=0.1\linewidth]{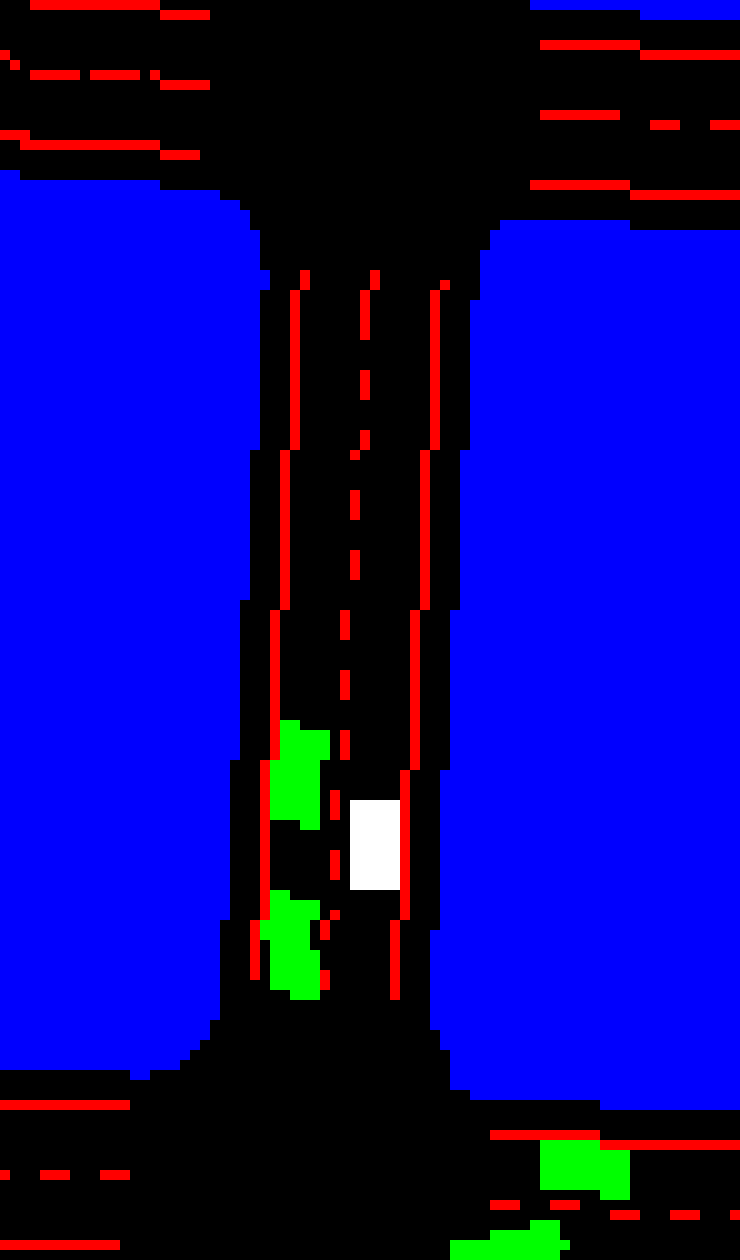}
		\caption{Ground truth sequence}
		\label{fig:fm_pred_truth_carla}
	\end{subfigure}
	\hfill
	\begin{subfigure}{\linewidth}
		\centering 
		\includegraphics[width=0.1\linewidth]{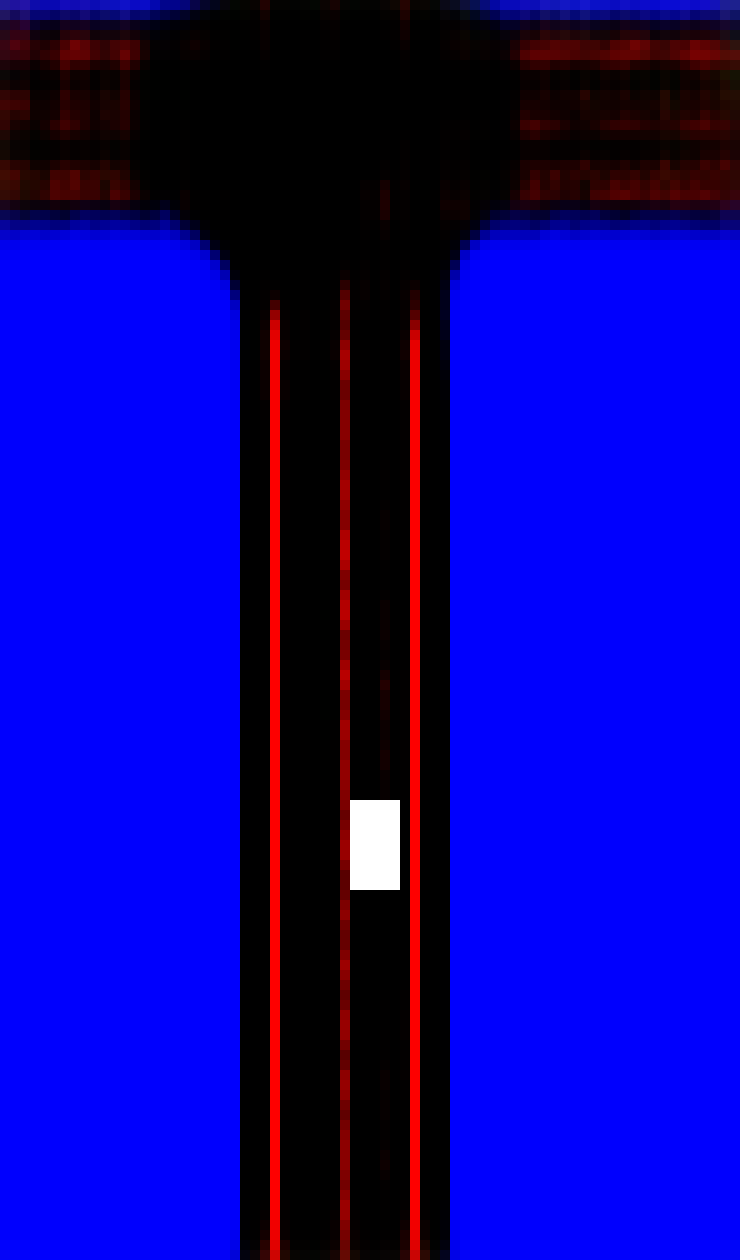}\hfill
		\includegraphics[width=0.1\linewidth]{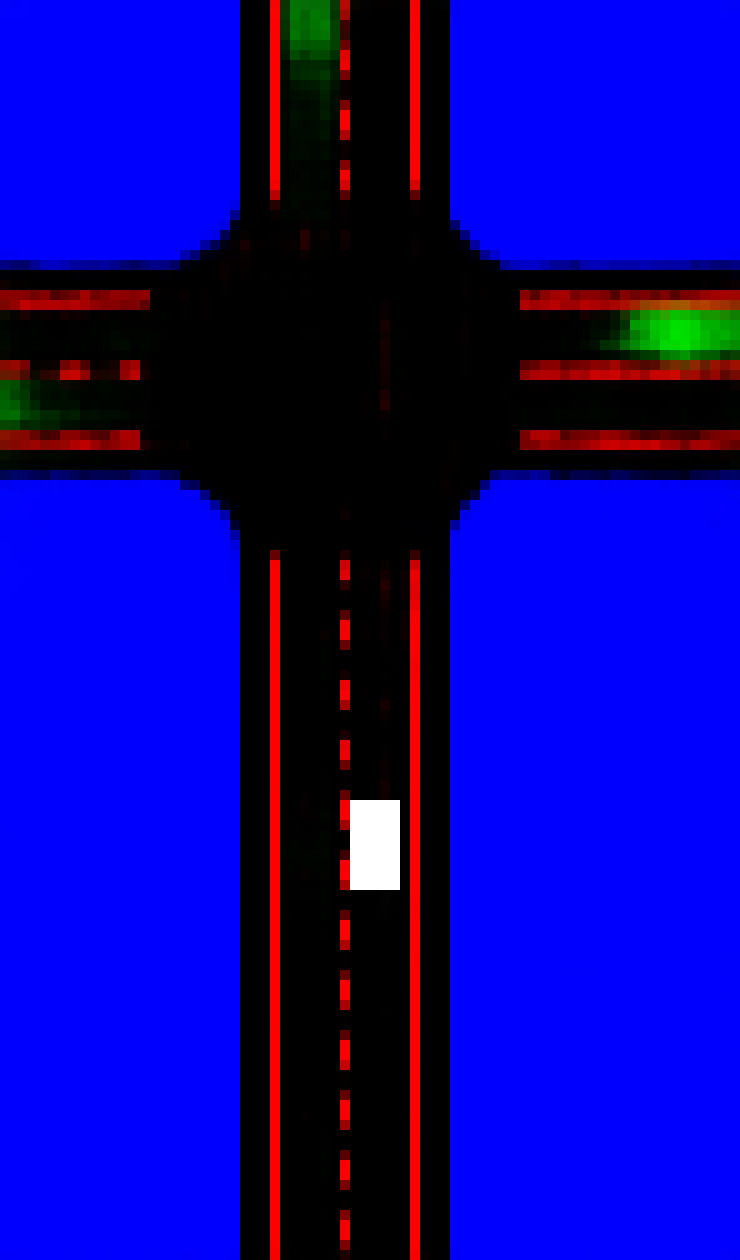}\hfill
		\includegraphics[width=0.1\linewidth]{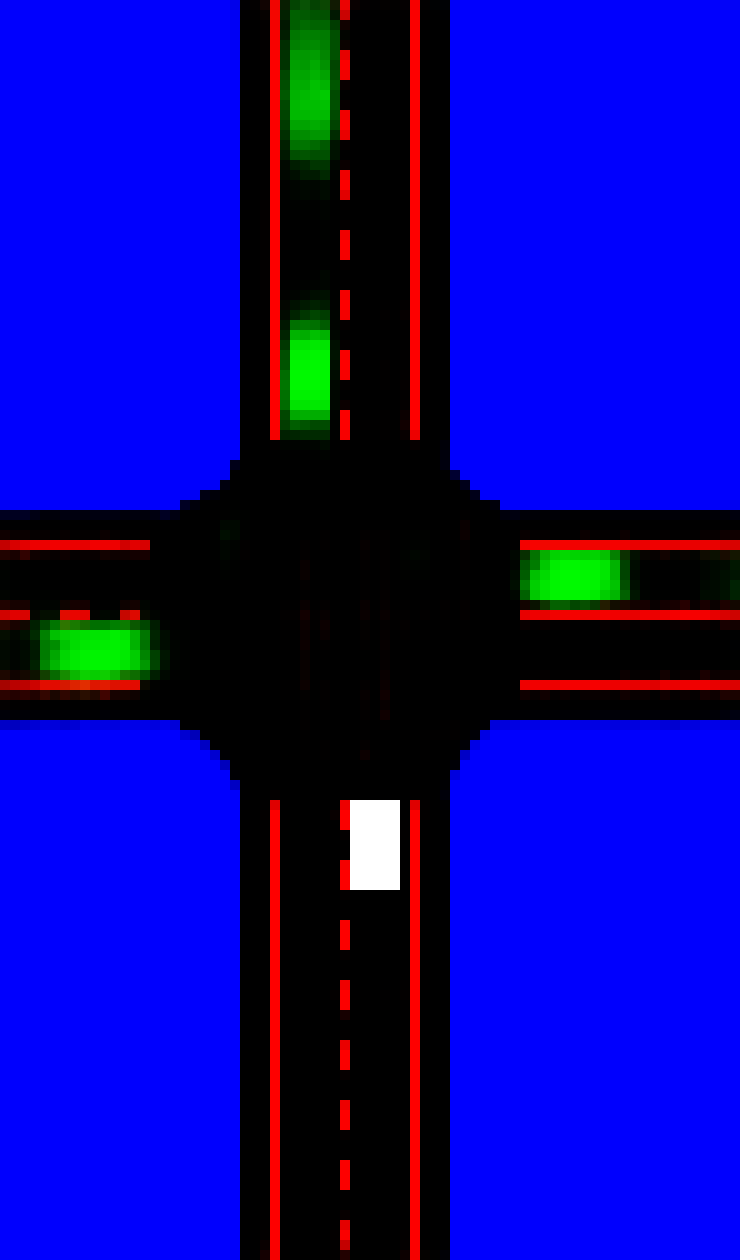}\hfill
		\includegraphics[width=0.1\linewidth]{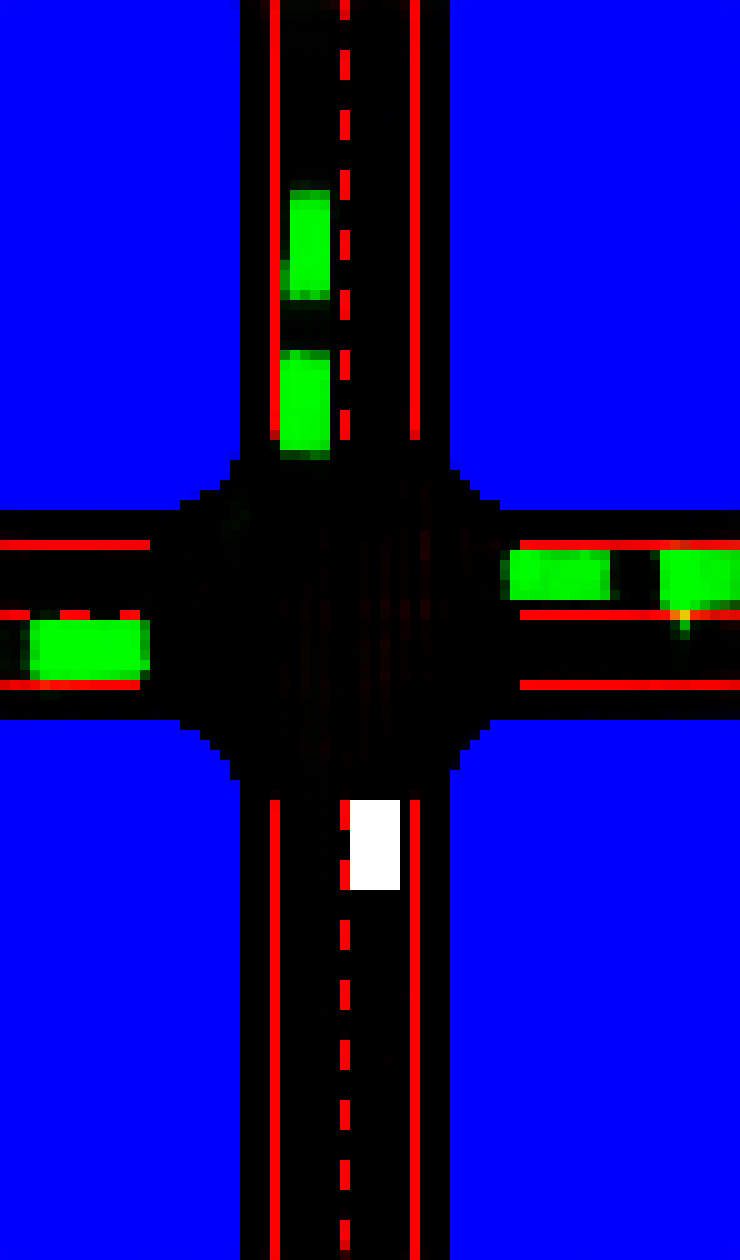}\hfill
		\includegraphics[width=0.1\linewidth]{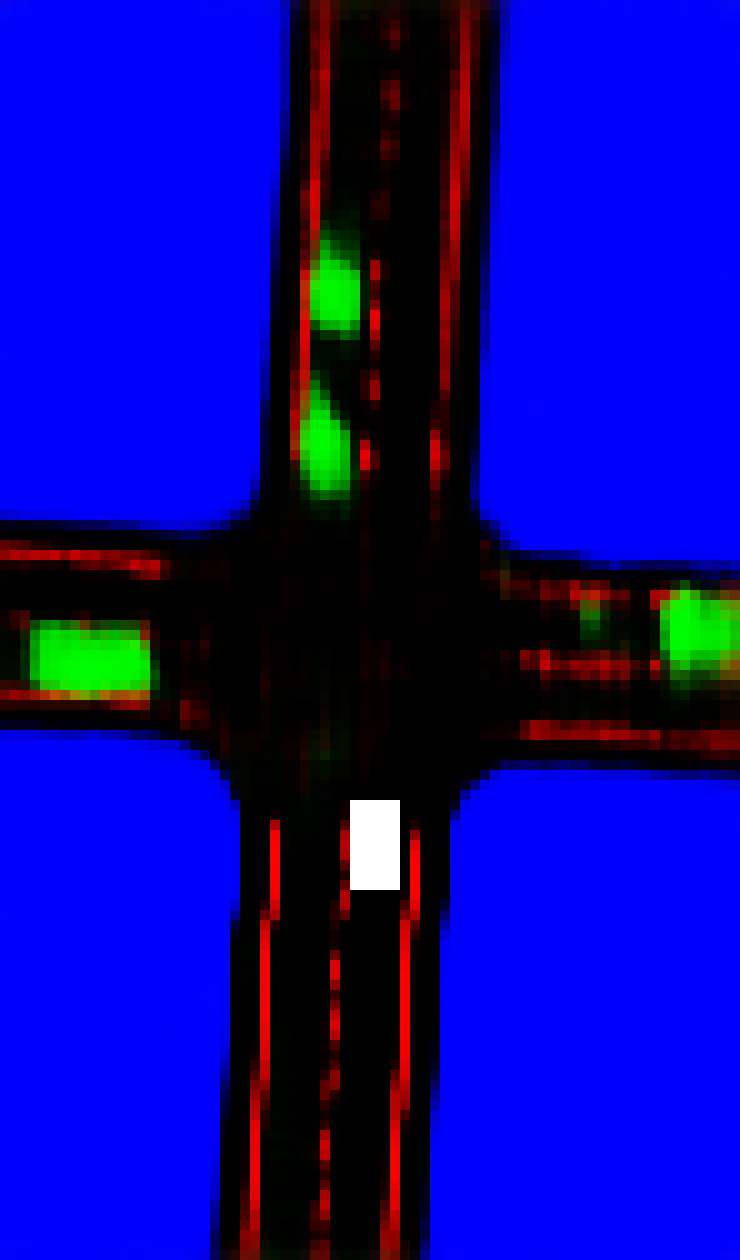}\hfill
		\includegraphics[width=0.1\linewidth]{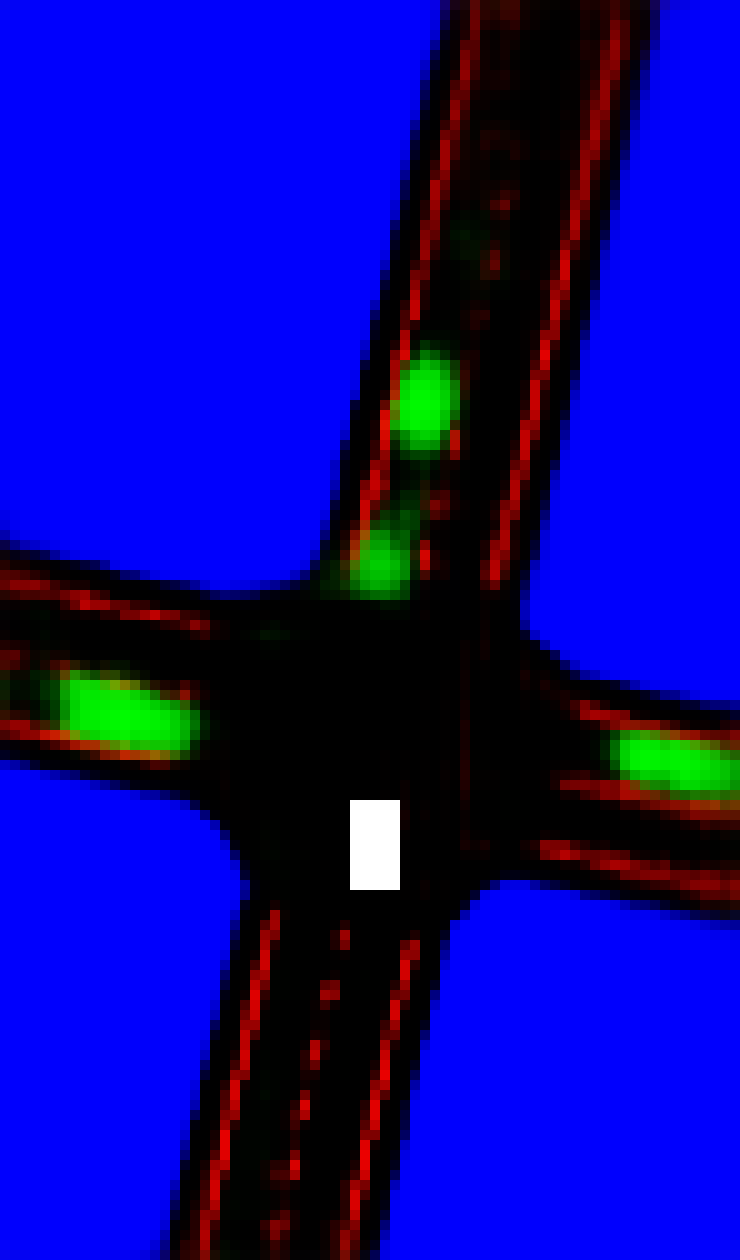}\hfill
		\includegraphics[width=0.1\linewidth]{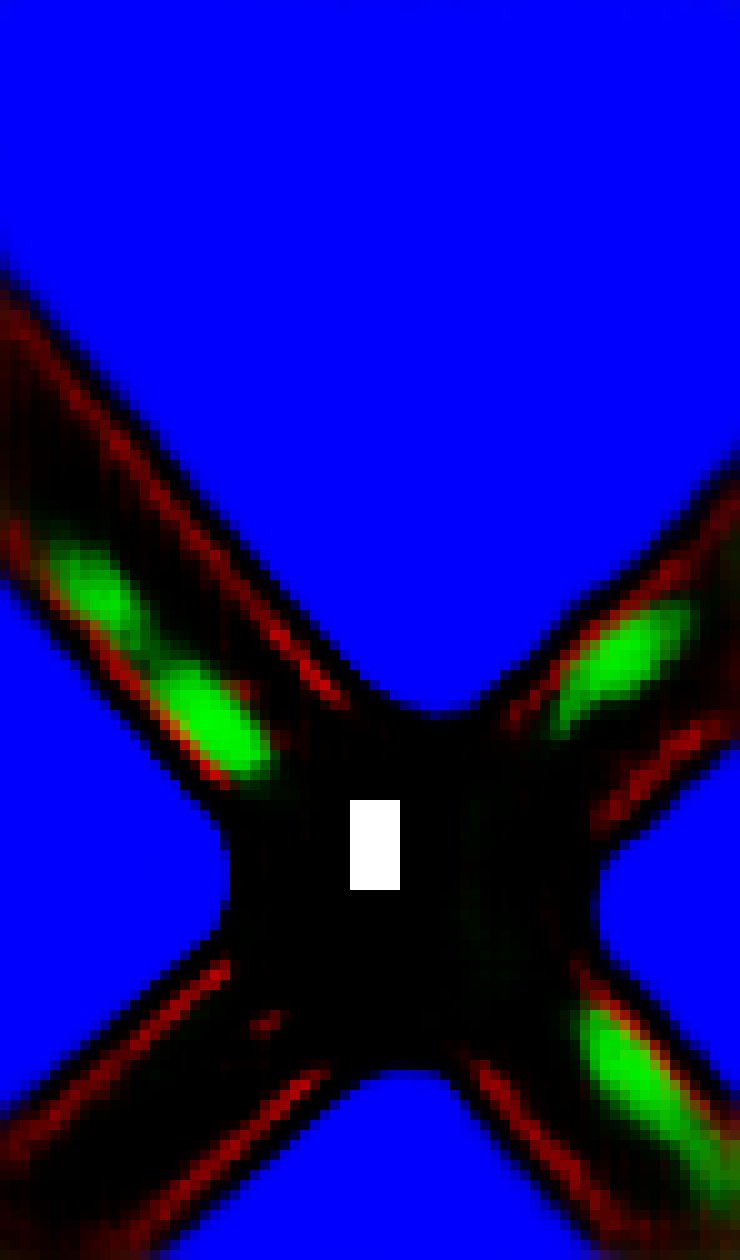}\hfill
		\includegraphics[width=0.1\linewidth]{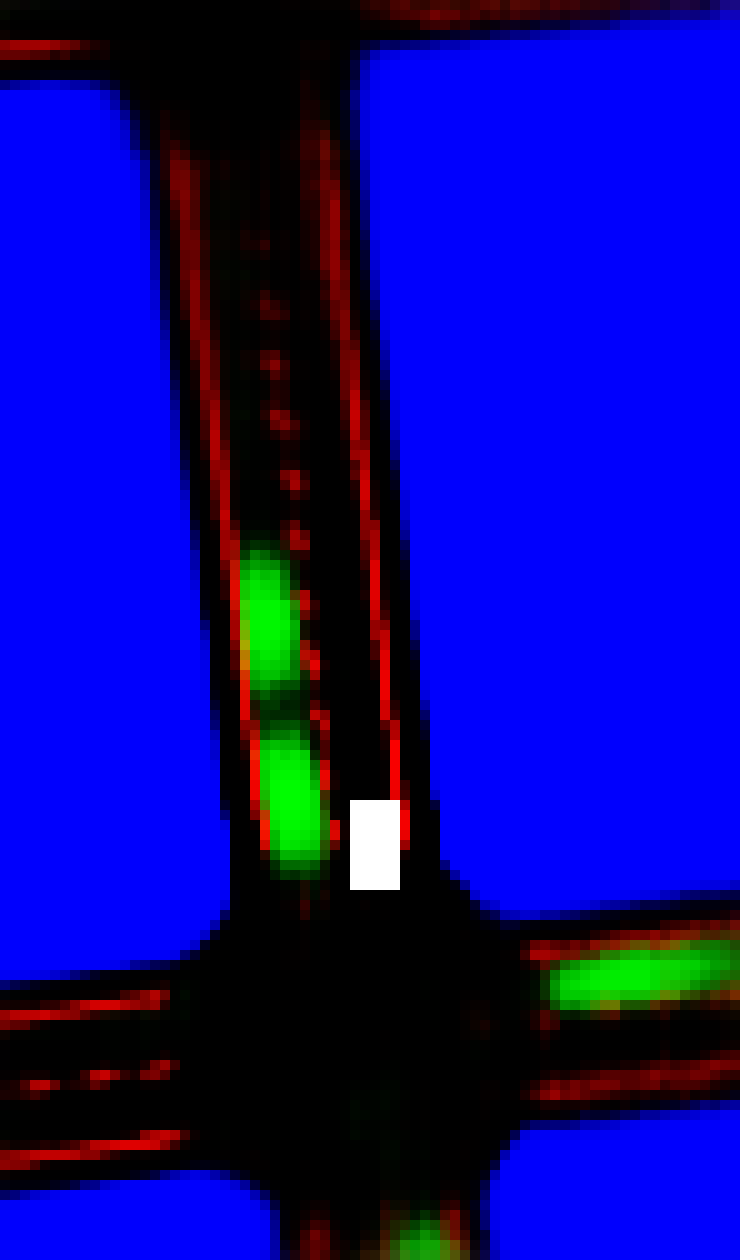}\hfill
		\includegraphics[width=0.1\linewidth]{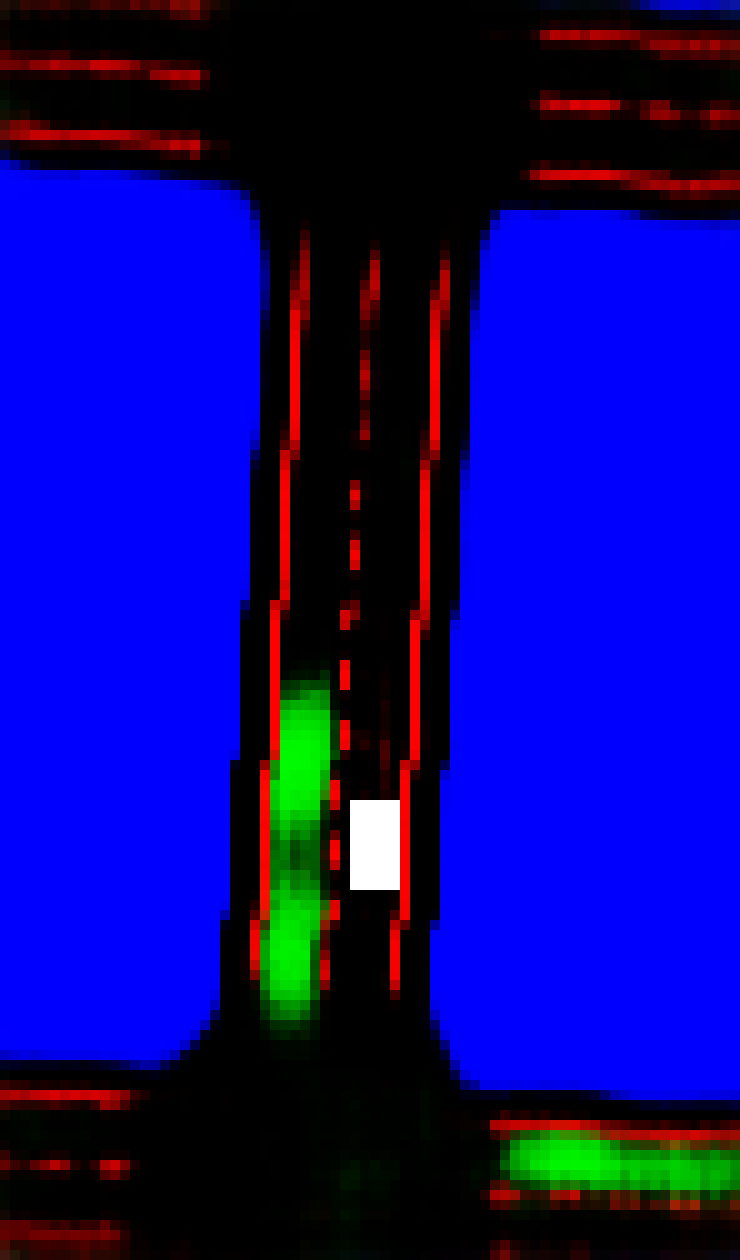}
		\caption{Predictions with deterministic model}
		\label{fig:fm_pred_det_carla}
	\end{subfigure}
	\begin{subfigure}{\linewidth}
		\centering
		\includegraphics[width=0.1\linewidth]{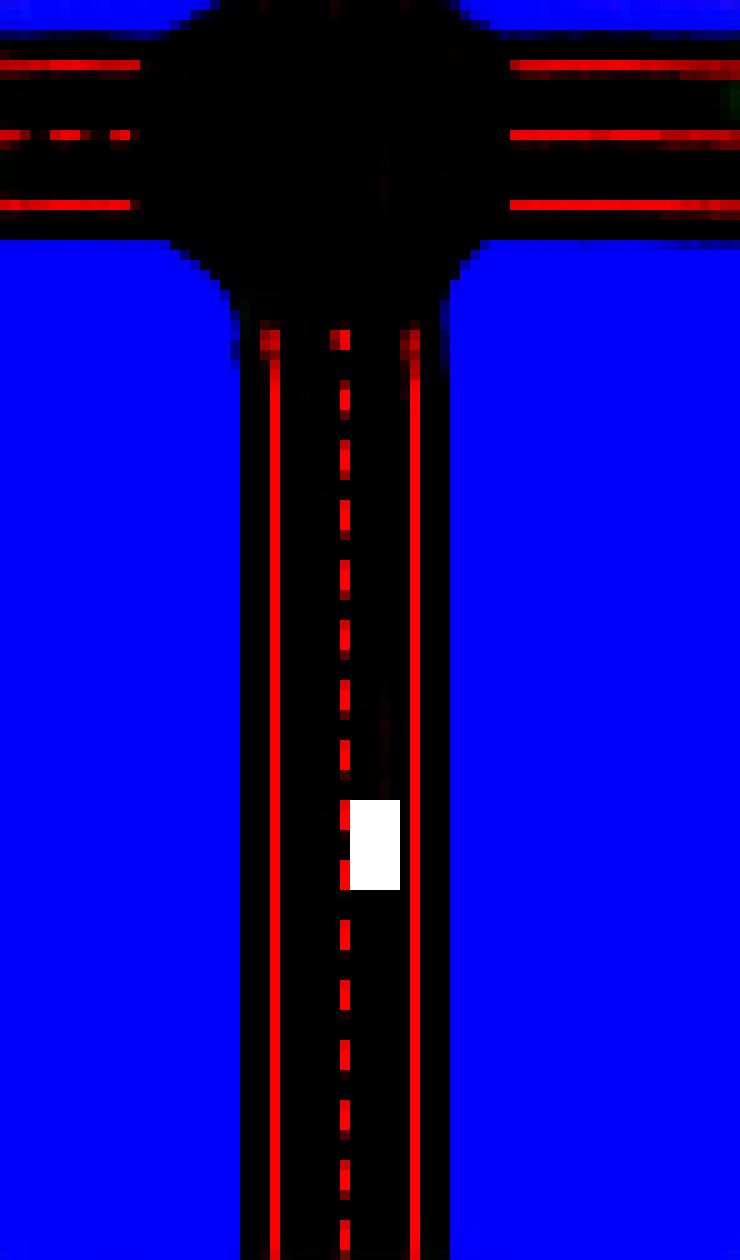}\hfill
		\includegraphics[width=0.1\linewidth]{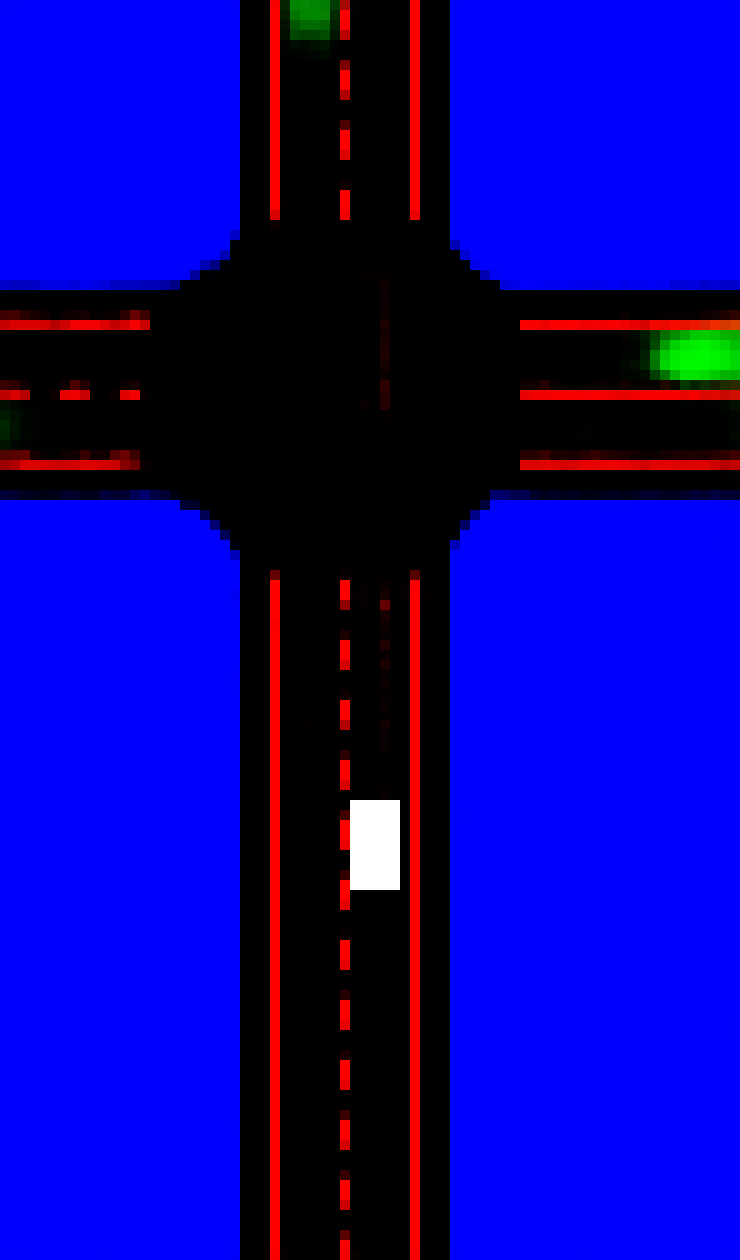}\hfill
		\includegraphics[width=0.1\linewidth]{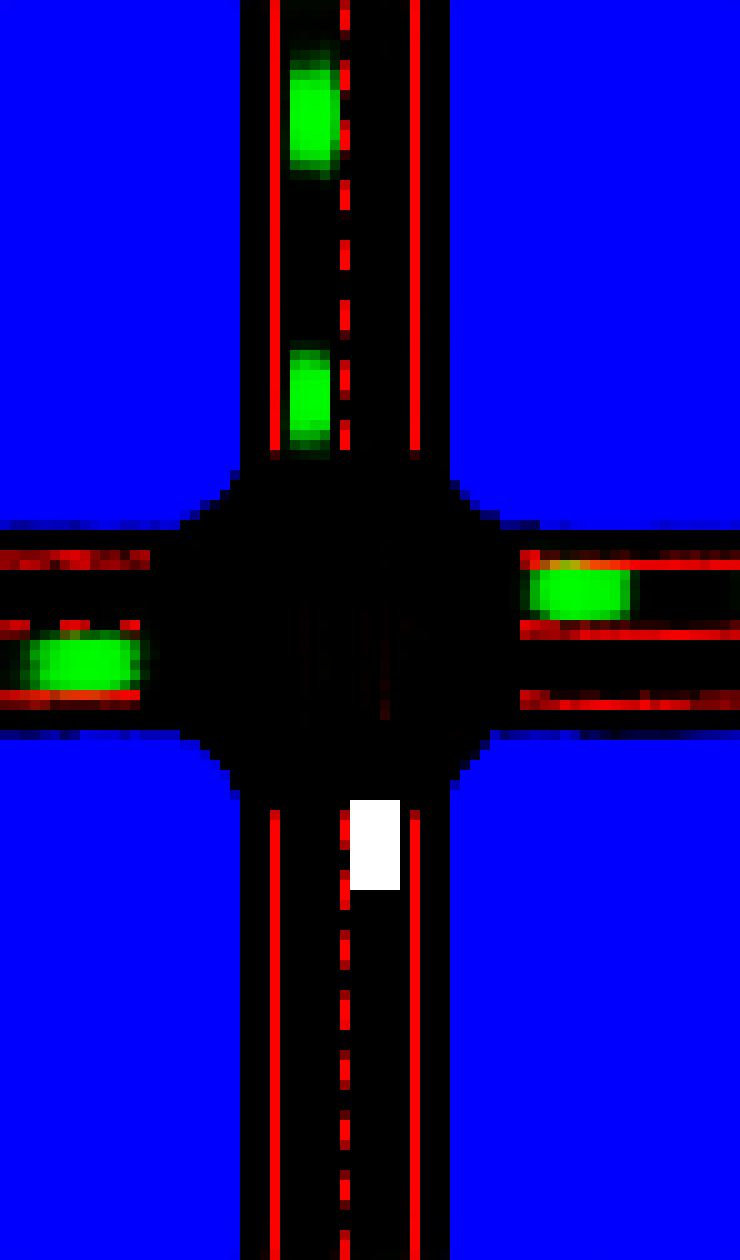}\hfill
		\includegraphics[width=0.1\linewidth]{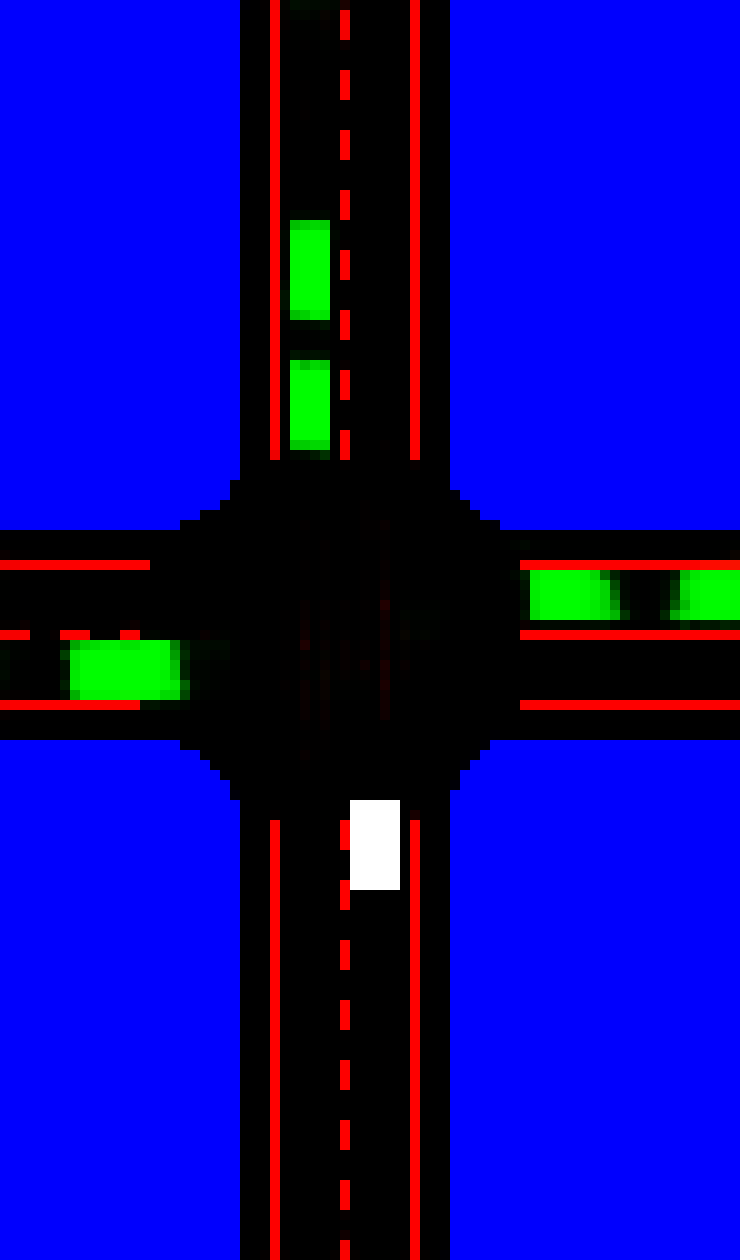}\hfill
		\includegraphics[width=0.1\linewidth]{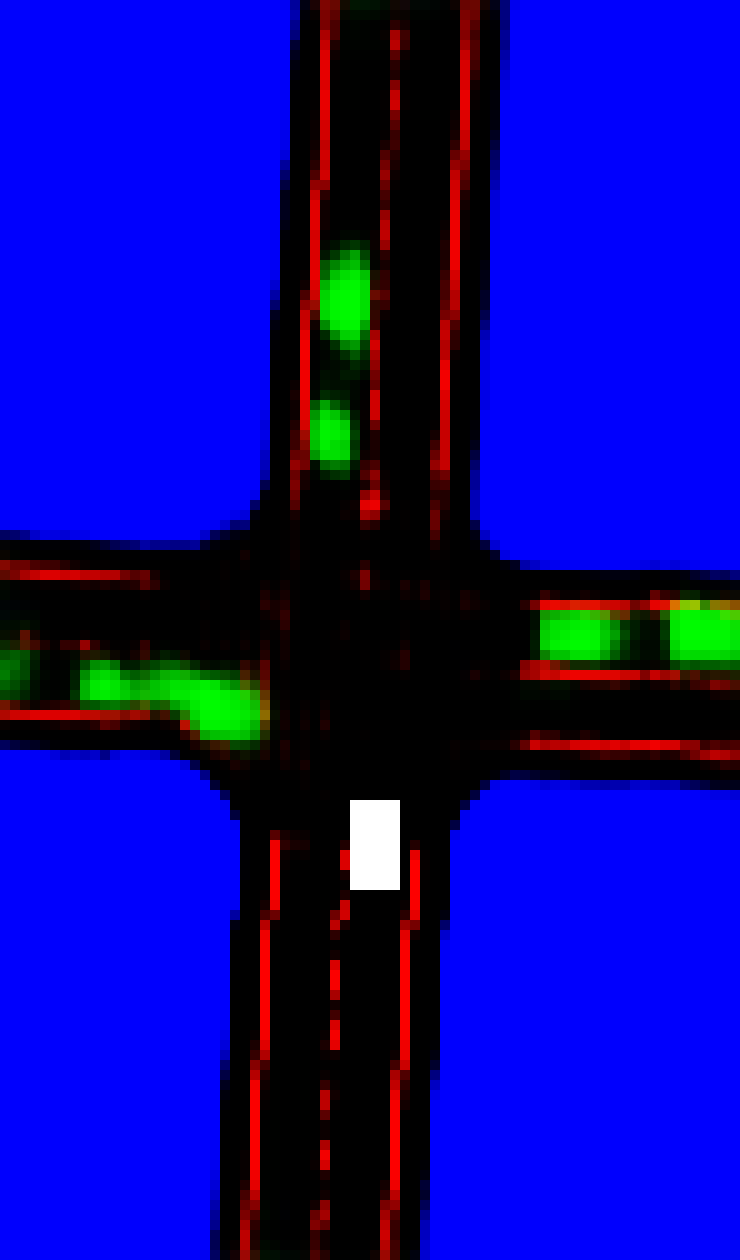}\hfill
		\includegraphics[width=0.1\linewidth]{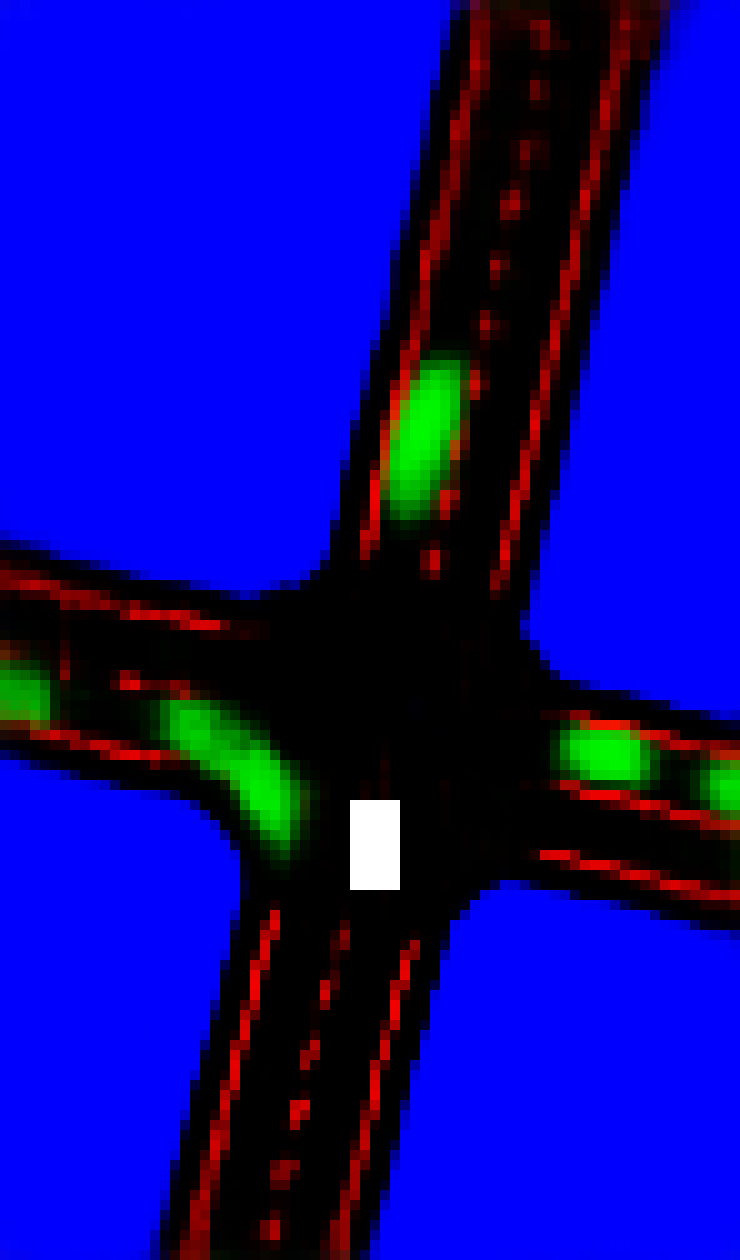}\hfill
		\includegraphics[width=0.1\linewidth]{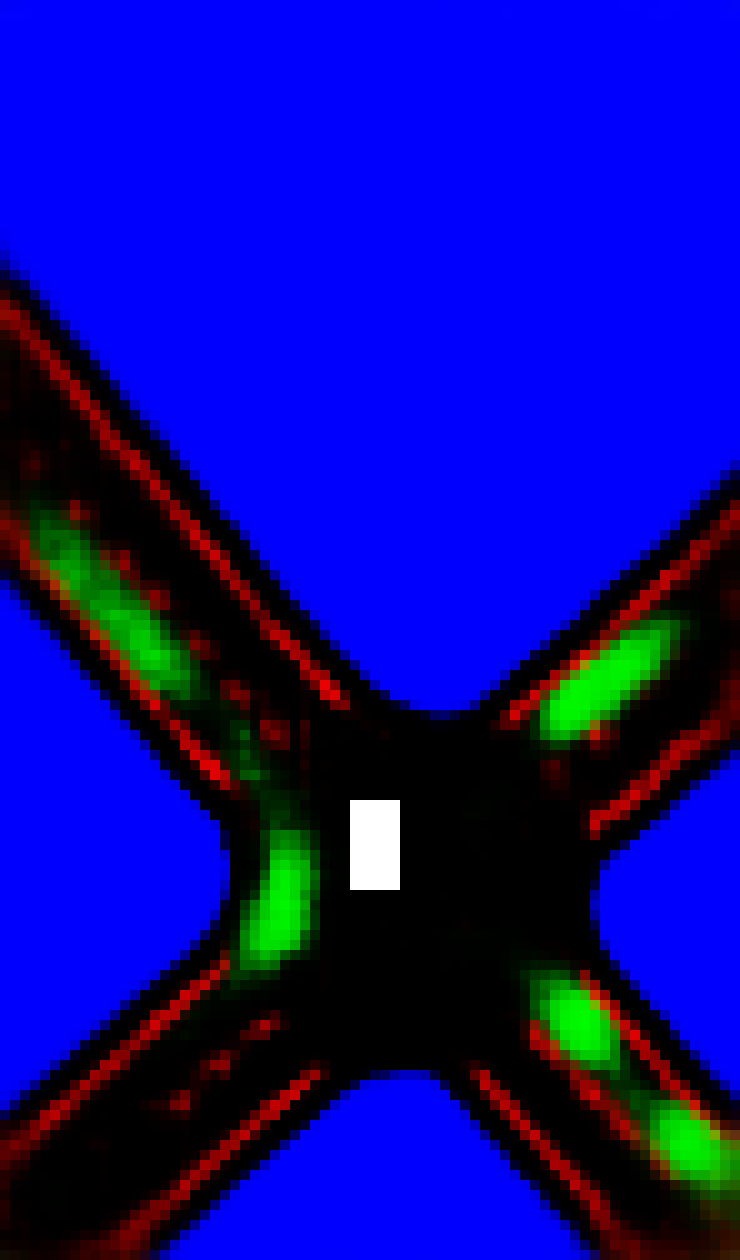}\hfill
		\includegraphics[width=0.1\linewidth]{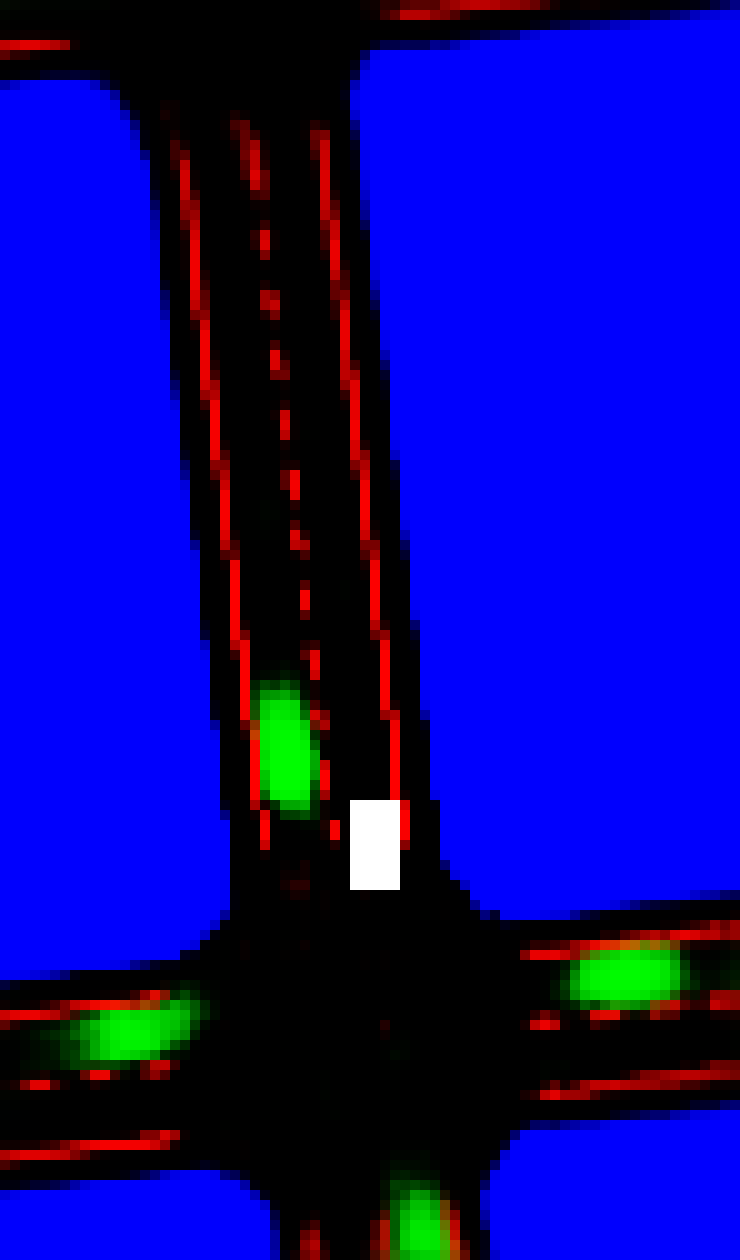}\hfill
		\includegraphics[width=0.1\linewidth]{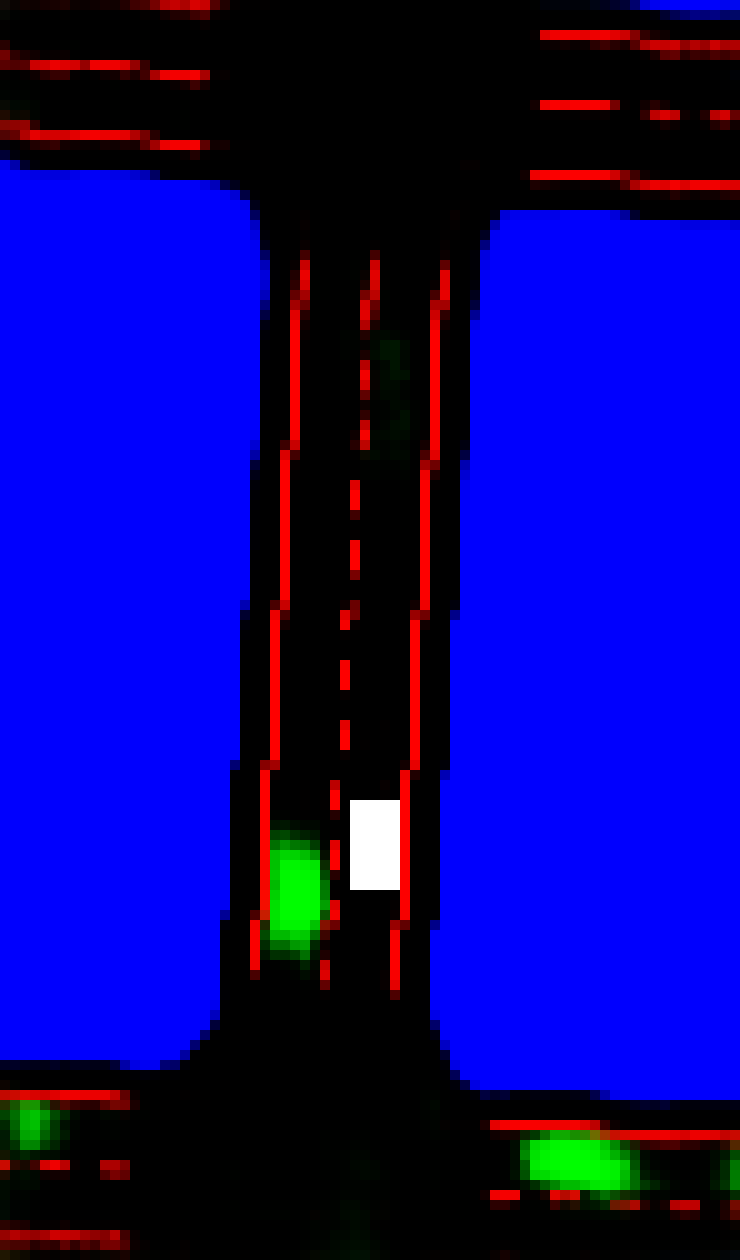}
		\caption{Predictions with stochastic model, $\textbf{z}$-sample 1}
		\label{fig:fm_pred_sto1_carla}
	\end{subfigure}
	\hfill
	\begin{subfigure}{\linewidth}
		\centering
		\includegraphics[width=0.1\linewidth]{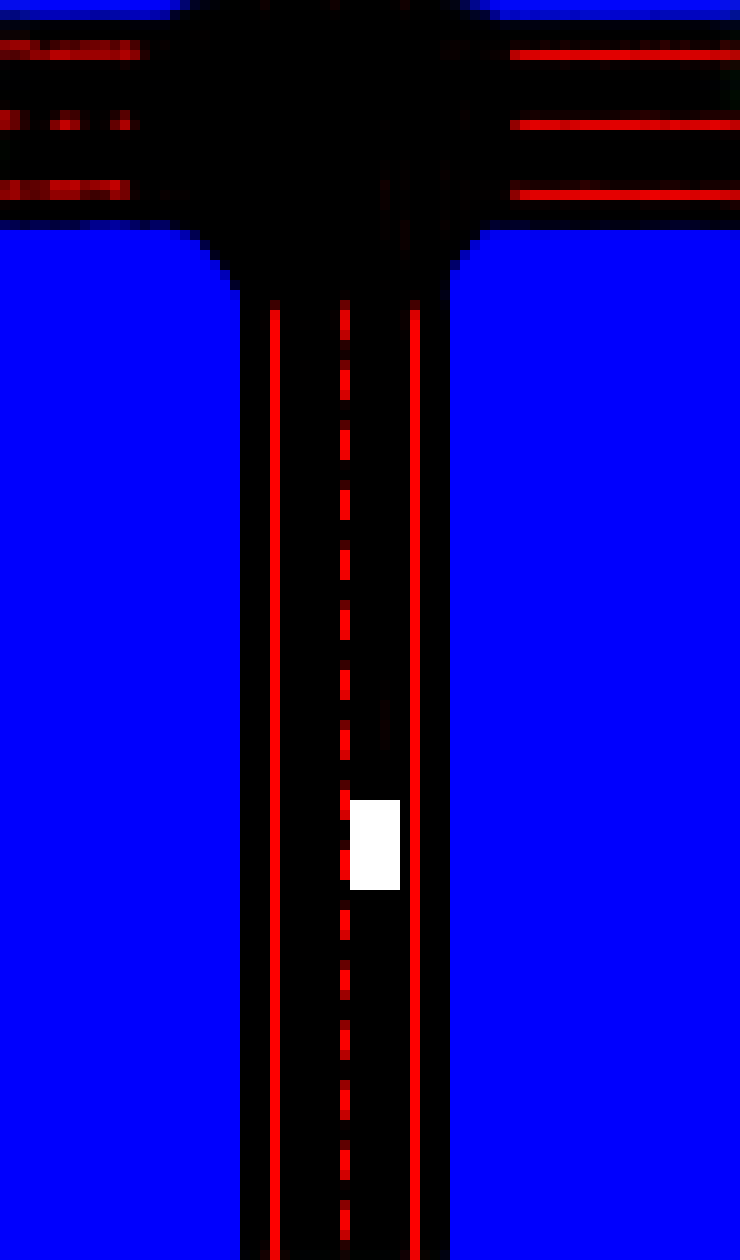}\hfill
		\includegraphics[width=0.1\linewidth]{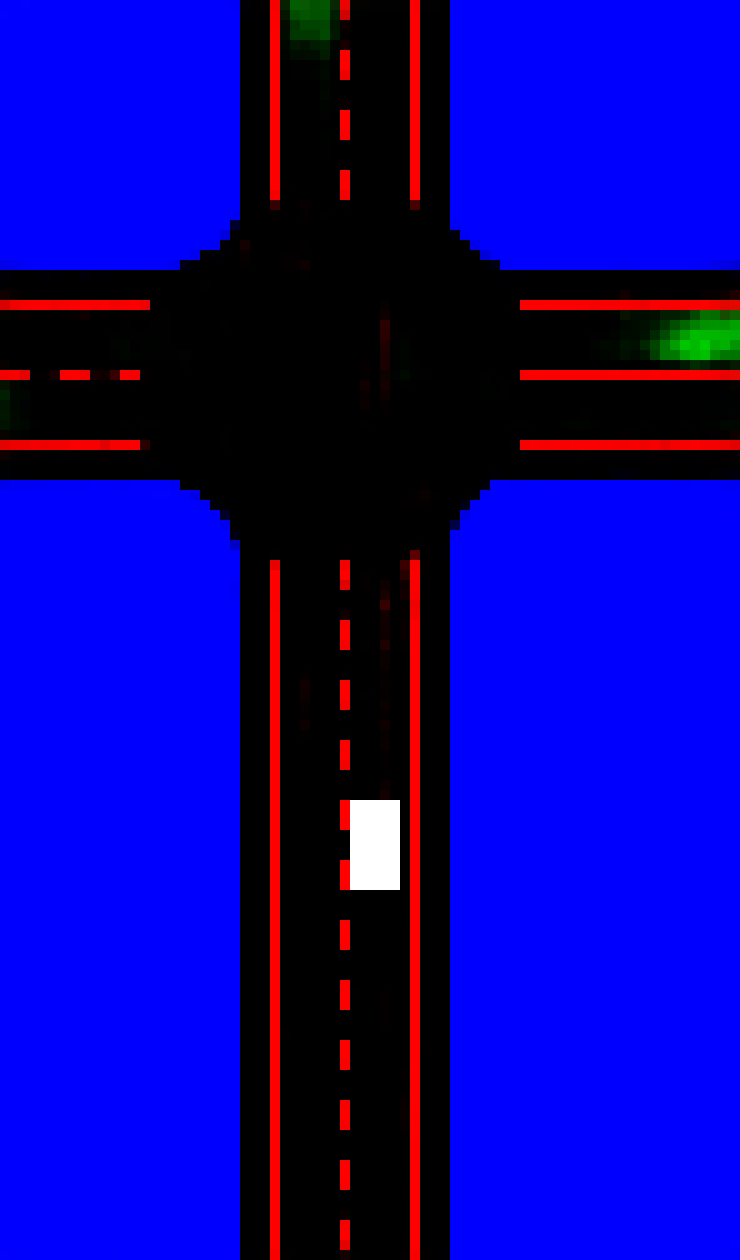}\hfill
		\includegraphics[width=0.1\linewidth]{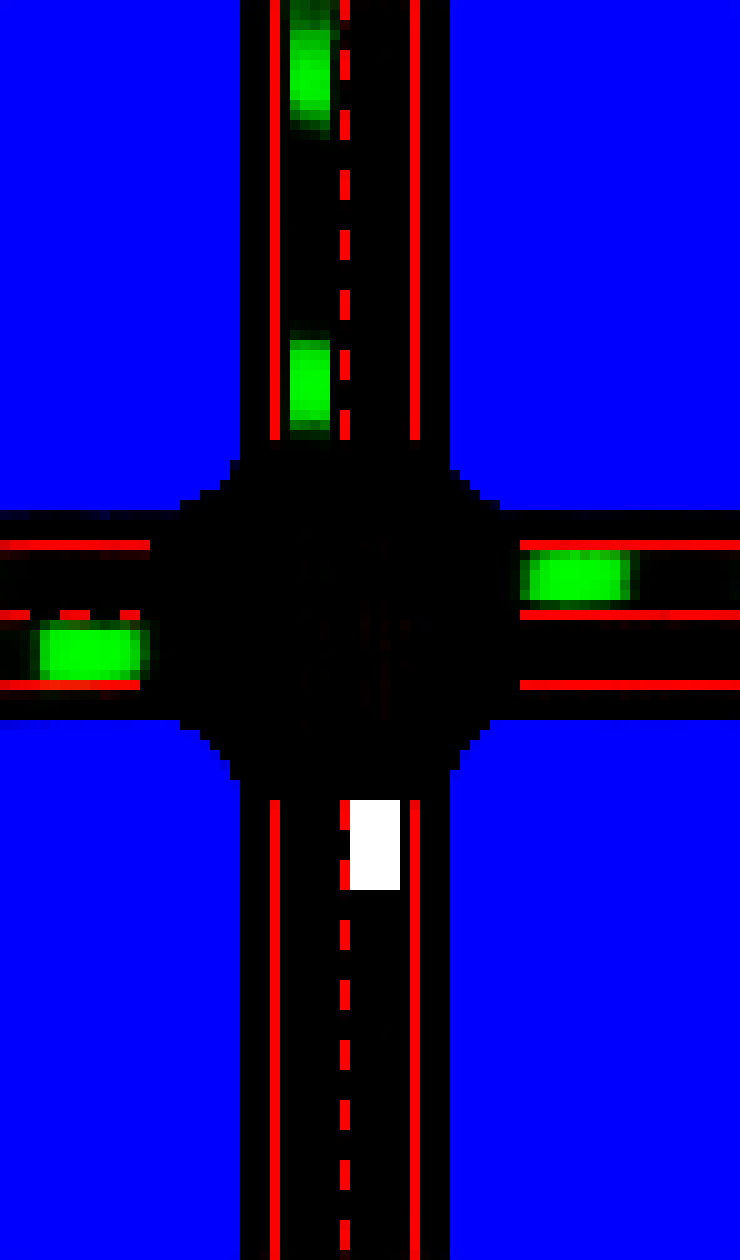}\hfill
		\includegraphics[width=0.1\linewidth]{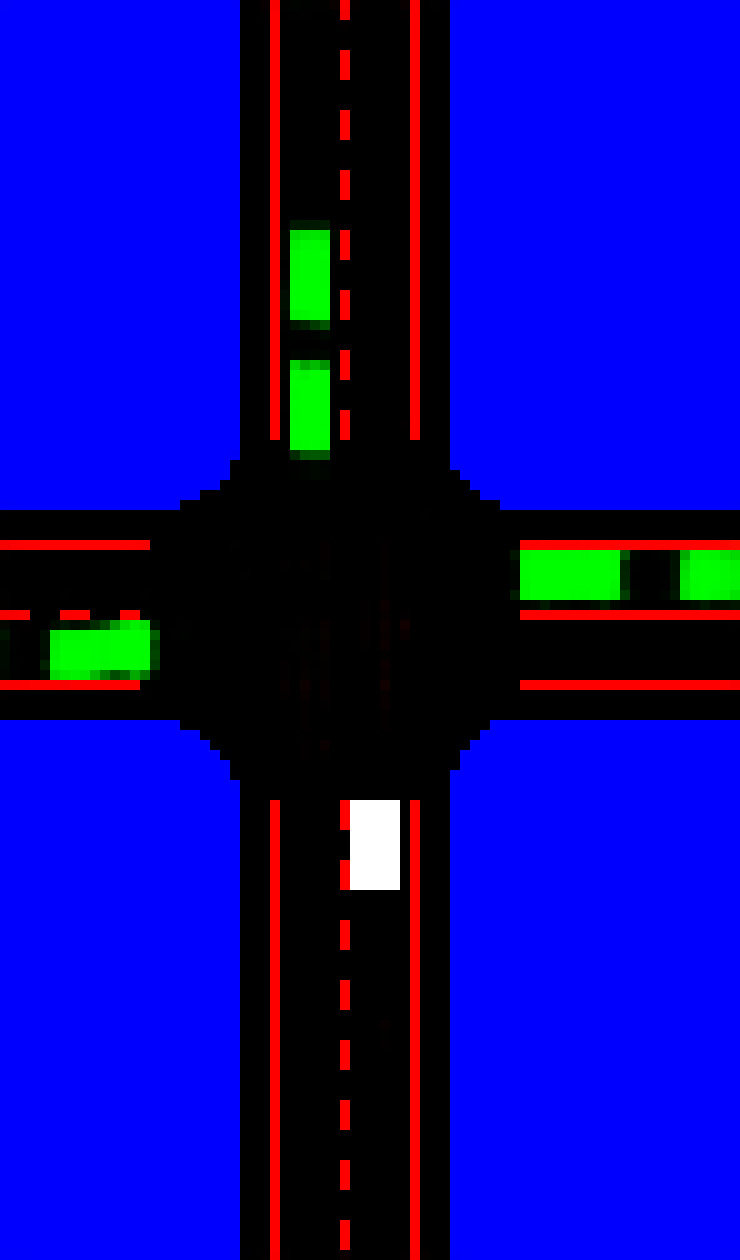}\hfill
		\includegraphics[width=0.1\linewidth]{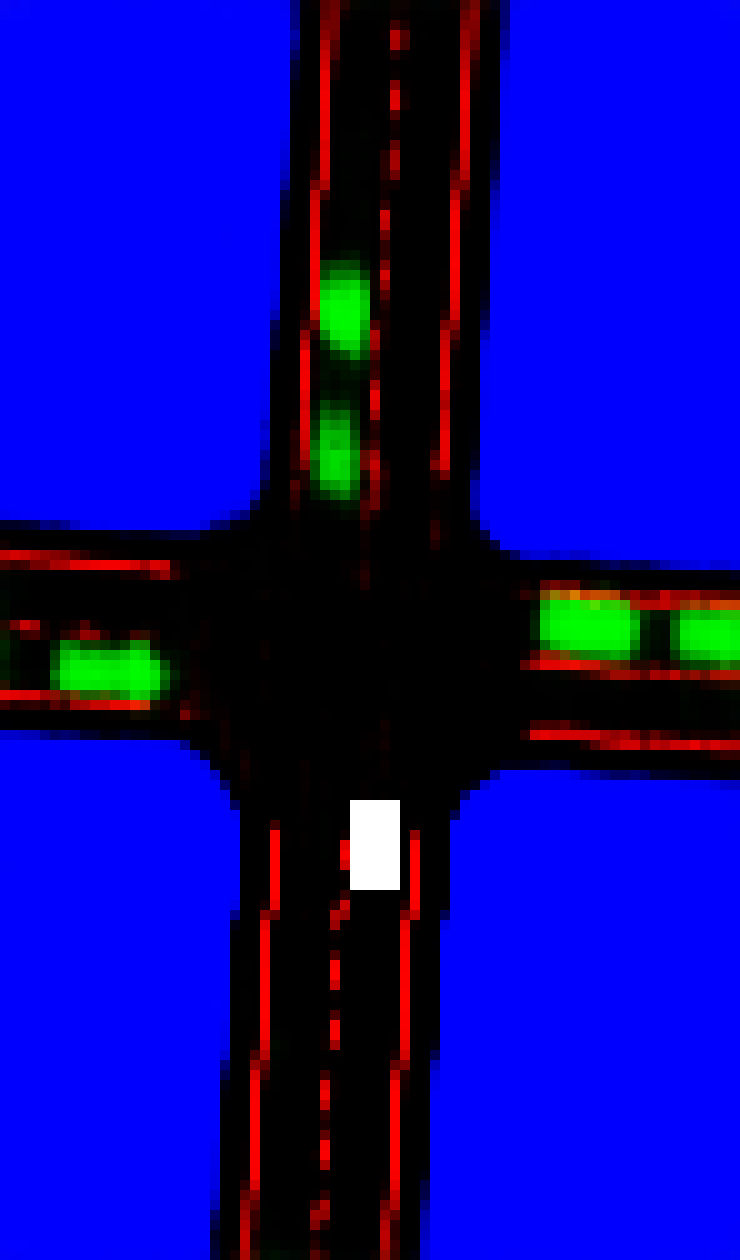}\hfill
		\includegraphics[width=0.1\linewidth]{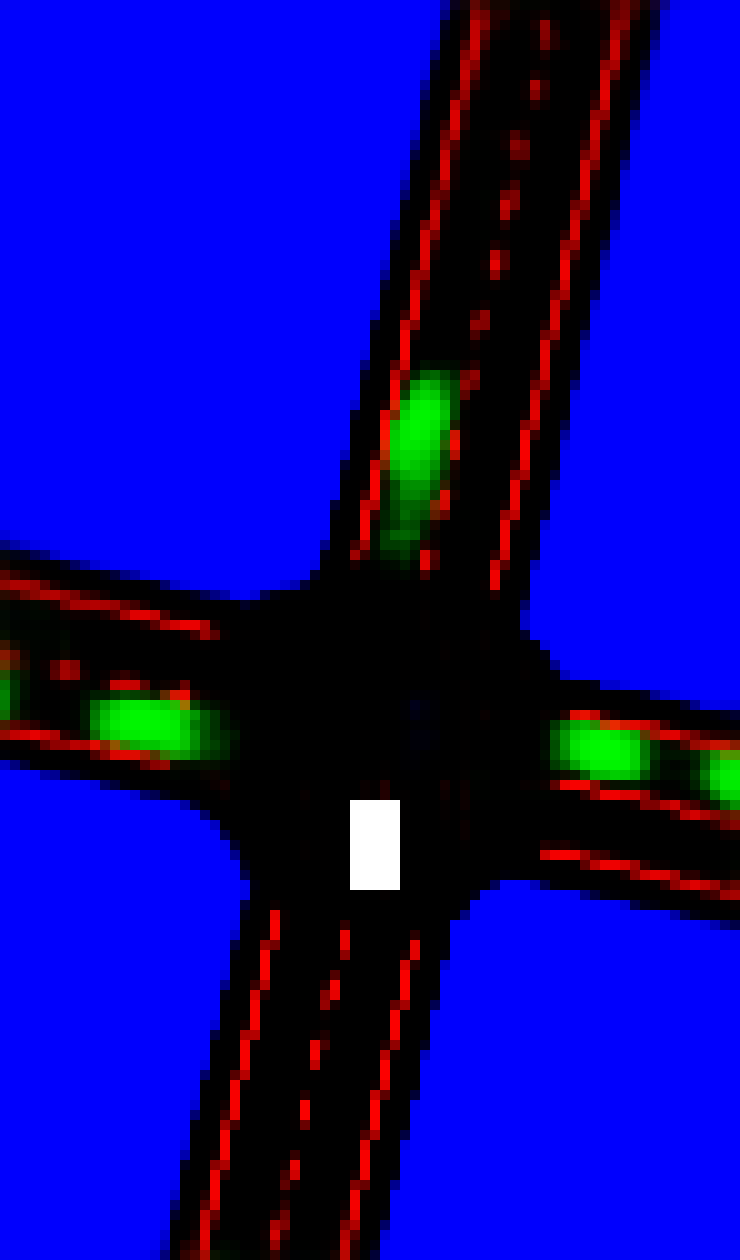}\hfill
		\includegraphics[width=0.1\linewidth]{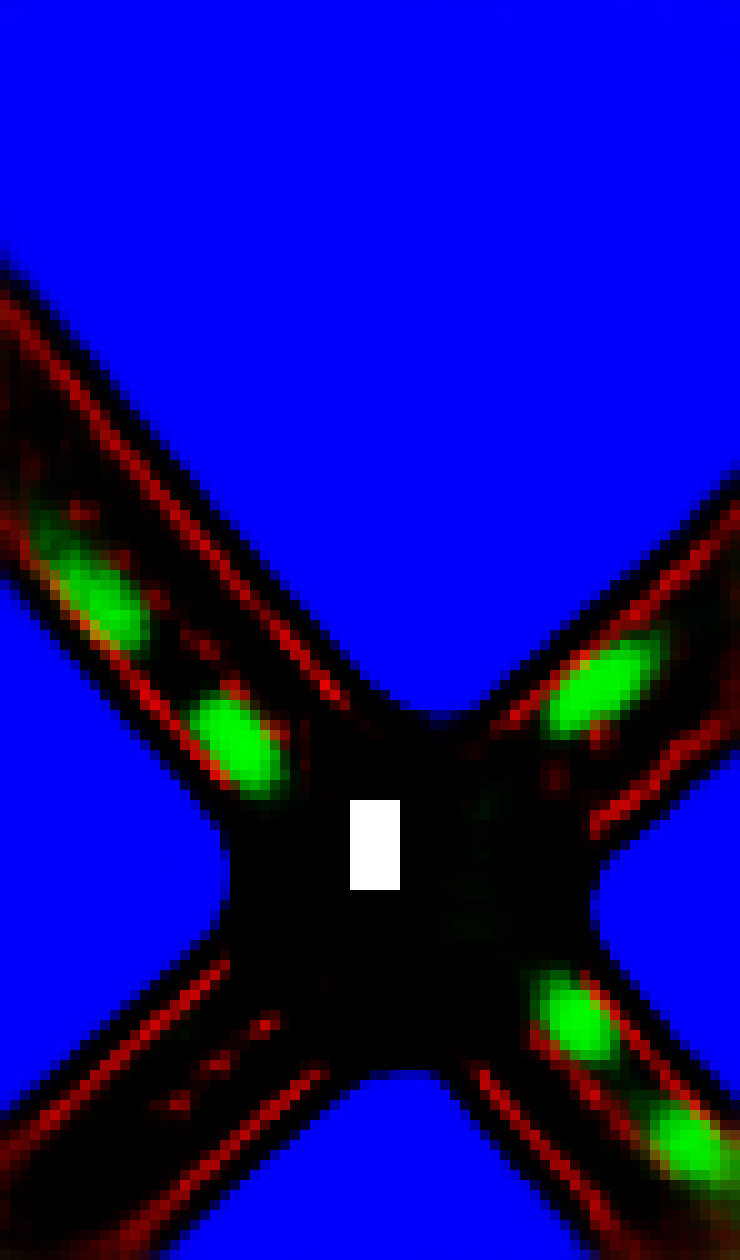}\hfill
		\includegraphics[width=0.1\linewidth]{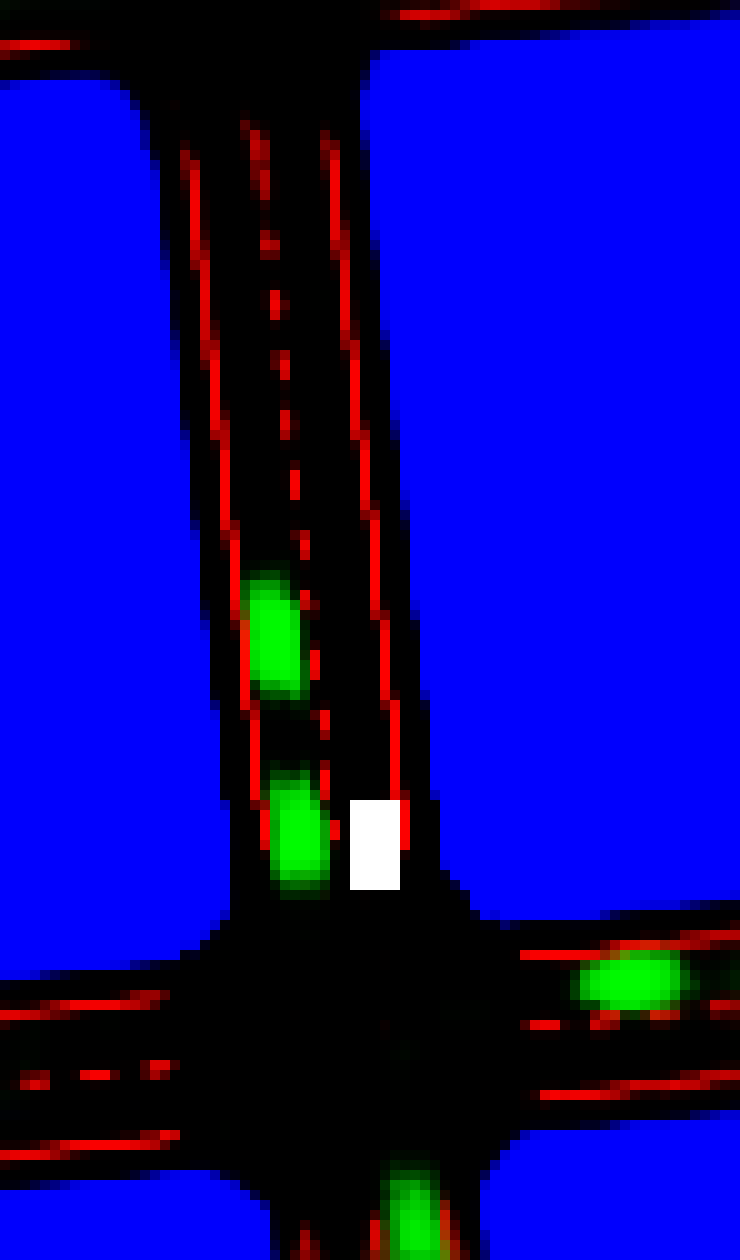}\hfill
		\includegraphics[width=0.1\linewidth]{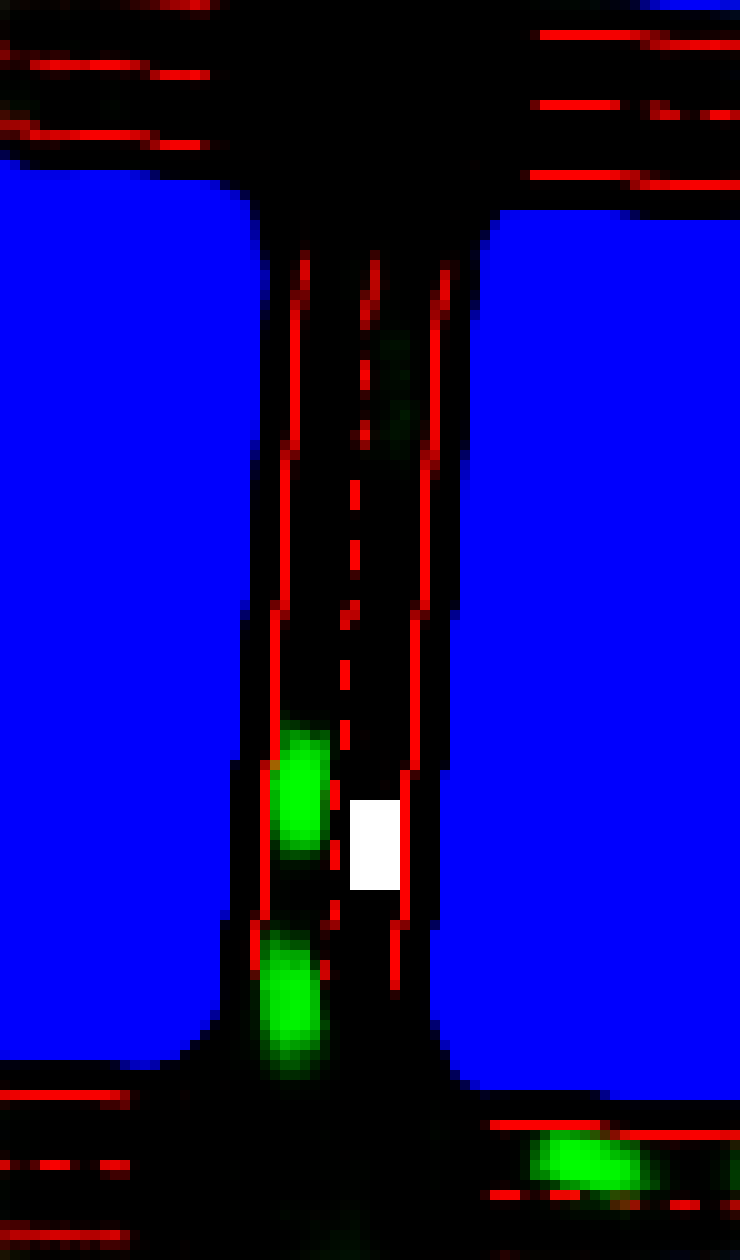}
		\caption{Predictions with stochastic model, $\textbf{z}$-sample 2}
		\label{fig:fm_pred_sto2_carla}
	\end{subfigure}	
	\caption{Video prediction results using the different models. (a) Ground truth sequence of the test dataset. (b) Predicted sequence made using the deterministic model (c),(d) Predicted sequence using the stochastic dynamics model. Sampling two different latent variables $\textbf{z}$ from the prior distribution results in two different predicted sequences.}
	\label{fig:fm_pred_carla}
\end{figure}

As the work of \cite{MPUR} already showed prediction results of a similar dynamics model in the NGSIM environment, this section only shows video predictions results from the CARLA experiments in Figure \ref{fig:fm_pred_carla}. The prediction horizon is $20\si{\second}$. We observe that the stochastic dynamics model can model different future outcomes over a long time horizon. The predictions of the deterministic model are more blurred (last column) as it is averaging over multiple futures. Both models have learned what the road geometry looks like after the intersection. Note that, the prediction results on NGSIM \cite{MPUR} are more blurred as different locations, vehicle types and more maneuvers occur during training.

\subsection{Runtime}
This work conducts runtime experiments using an Nvidia Titan X and an Intel i5-3550@3.3 GHz. In its current version, the implementations of MBOP and UMBRELLA are not real-time capable using the models described in Appendix \ref{Model Architecture}, as shown in Figure \ref{fig:time_ef_2}. However, the runtime could be improved as the current implementation is not optimized for performance. Moreover, both methods encode and decode high-dimensional images using CNN's leading to a high computation cost. Note, that high computation times also occur when using MBOP with the proposed context image-based network. Hence, future work should focus on using more efficient networks. For example, the observation space could be represented by a graph \citep{VectorNet, LaneGCN} instead of an image. Then using graph neural networks \citep{GraphNeuralNetwork} would lead to a more efficient implementation regarding computation time and memory consumption. 

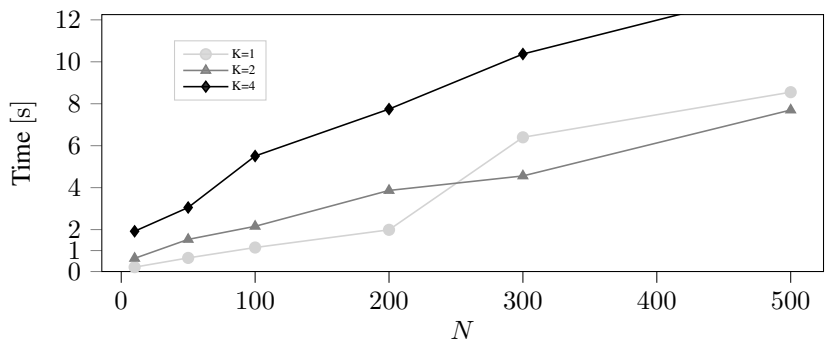
\begin{figure}[h]
	\centering
\begin{tikzpicture}

\begin{axis}[
width=0.8\textwidth,
height=5cm,
legend cell align={left},
legend style={fill opacity=0.8, draw opacity=1, text opacity=1, draw=white!80!black, at={(0.1,0.9)},anchor=north west, nodes={scale=0.5, transform shape}},
tick align=outside,
tick pos=left,
x grid style={white!69.0196078431373!black},
xlabel={$N$},
xmin=-14.5, xmax=524.5,
xtick style={color=black},
ytick={0.0,1,2,4,6,8,10,12},
y grid style={white!69.0196078431373!black},
ylabel={Time $\left[\si{\second}\right]$ },
ylabel near ticks,
ymin=0.0, ymax=12.25,
]
\addplot [semithick, white!82.7450980392157!black, mark=*, mark size=2, mark options={solid}]
table {%
10 0.208836589900541
50 0.649178343147084
100 1.14501961659829
200 1.98940731982128
300 6.39955783400696
500 8.55134053983186
};
\addlegendentry{K=1}
\addplot [semithick, white!50.1960784313725!black, mark=triangle*, mark size=2, mark options={solid}]
table {%
10 0.628784513707637
50 1.53249512270989
100 2.15585246204205
200 3.8656199634622
300 4.55863939835145
500 7.70043479093331
};
\addlegendentry{K=2}
\addplot [semithick, black, mark=diamond*, mark size=2, mark options={solid}]
table {%
10 1.92022471773259
50 3.05286800097622
100 5.50746065492083
200 7.74502042682828
300 10.3706190106661
500 13.5952461005972
};
\addlegendentry{K=4}
\end{axis}

\end{tikzpicture}
	\caption{Average computation time of UMBRELLA as function of $N$ and $K$ with a horizon of $H=30$.}
	\label{fig:time_ef_2}
\end{figure}
\FloatBarrier
\subsection{Hyperparameter Sensitivity}

\begin{table}[!t]
	\centering
	\caption{Performance of UMBRELLA in the NGSIM environment for varying parameter combinations. Only one hyperparameter is changed at a time and the rest of the configuration from Table \ref{tab:planning_param}}.
	\begin{tabular}{lcc}
		\toprule
		Parameter & SR & MD \\
		\midrule
		$K = 1$ & $0.43$ & $266.71$ \\
		$K = 4$ & $0.43$ & $284.46$ \\
		\midrule
		$H = 25$ & $0.57$ & $240.57$ \\
		$H = 35$ & $0.60$ & $293.37$ \\
		$H = 40$ & $0.50$ & $280.30$ \\
		\midrule
		$N = 10$ & $0.20$ & $221.72$ \\
		$N = 50$ & $0.47$ & $290.65$ \\
		$N = 100$ & $0.50$ & $281.62$\\
		$N = 200$ & $0.53$ & $297.44$\\
		$N = 500$ & $0.50$ & $299.71$\\
		\midrule
		$\sigma^2 = 0.3$ & $0.43$ & $273.21$ \\
		$\sigma^2 = 0.5$ & $0.50$ & $282.07$ \\
		$\sigma^2 = 0.7$ & $0.53$ & $296.21$ \\
		$\sigma^2 = 1.0$ & $0.53$ & $283.80$ \\
		$\sigma^2 = 1.5$ & $0.50$ & $303.52$ \\
		$\sigma^2 = 3.0$ & $0.27$ & $202.44$ \\
		\midrule
		$\beta = 0.0$ & $0.03$ & $116.06$ \\
		$\beta = 0.1$ & $0.07$ & $128.12$ \\
		$\beta = 0.3$ & $0.13$ & $223.59$ \\
		$\beta = 0.5$ & $0.53$ & $307.63$ \\
		$\beta = 0.7$ & $0.40$ & $284.99$ \\
		$\beta = 0.9$ & $0.20$ & $202.57$ \\
		\midrule
		$\kappa = 0.01$ & $0.10$ & $182.77$\\
		$\kappa = 0.1$ & $0.40$ & $272.43$ \\
		$\kappa = 0.3$ & $0.53$ & $339.97$ \\
		$\kappa = 0.5$ & $0.60$ & $320.06$ \\
		$\kappa = 1.0$ & $0.47$ & $292.17$ \\
		$\kappa = 1.5$ & $0.57$ & $287.37$ \\
		$\kappa = 3.0$ & $0.20$ & $166.66$ \\
		\bottomrule

	\end{tabular}
	\label{tab:param_eval_NGSIM}
\end{table}

Additional experiments to research the sensitivity to hyperparameters are conducted. Table \ref{tab:param_eval_NGSIM} shows the performance of UMBRELLA in the NGSIM environment for varying individual hyperparameters based on the configuration from Table \ref{tab:planning_param}. 

While the most sensitive hyperparameter is $\beta$, the algorithm remains stable for different planning horizons $H$. $\beta$ can be interpreted as a way to introduce momentum to the algorithm. A too low value makes the actions independent of previous actions, leading to behaviors with high jerk. However, a too high value makes the SDV less reactive. Future, work could also investigate the effect of a context dependent $\beta$ value. As the work of \cite{MBOP}, UMBRELLA performance mostly degrades near extreme values. We note that the prior work evaluates the sensitivity based on rewards. However as discussed in Section \ref{Limitations} and the work \cite{Rewardmisdesign}, this often does not necessarily correlate with a good driving performance.
\end{document}